\documentclass{article}

     \usepackage[preprint,nonatbib]{neurips_2020}
\usepackage[numbers,sort&compress]{natbib}
\usepackage[utf8]{inputenc} % allow utf-8 input
\usepackage[T1]{fontenc}    % use 8-bit T1 fonts
\usepackage{hyperref}       % hyperlinks
\usepackage{url}            % simple URL typesetting
\usepackage{booktabs}       % professional-quality tables
\usepackage{amsfonts}       % blackboard math symbols
\usepackage{nicefrac}       % compact symbols for 1/2, etc.
\usepackage{microtype}      % microtypography
\usepackage{arydshln}
\usepackage{threeparttable}
\usepackage{graphicx} % Required for inserting images
\usepackage{amsmath}
\usepackage{amssymb}
\usepackage{booktabs}
\usepackage{caption}
\usepackage{graphicx}
\usepackage{multirow}
\usepackage{hyperref}
\usepackage{algorithm}
\usepackage{algorithmic}
\usepackage{enumitem}
\usepackage{subfigure}
\usepackage{float}
\usepackage{multirow}
\usepackage{wrapfig}
\usepackage{tcolorbox}
\PassOptionsToPackage{dvipsnames}{xcolor}
\usepackage{xcolor}
\definecolor{typegreen}{HTML}{7BC8A4}  % Mint green for Agreement
\definecolor{typeblue}{HTML}{3498db}   % Blue for Type 2
\definecolor{typepink}{HTML}{E36387}   % Pink/rose for Type 1 and Both
\definecolor{typepurple}{HTML}{C900C5}   % Pink/rose for Type 1 and Both

\title{KUDEX: A Zero-Shot Decompositional Framework for Logical Analysis in Natural Language}

\author{%
  Zixu Zhao, Tianhang Zhang, Tong He, Zheng Zhang \\
  AWS \\
  \texttt{zhaozixu@amazon.com} \\
}

\begin{document}

\maketitle

\begin{abstract}
%Traditional methods for analyzing the relationship between statements, such as Natural Language Inference (NLI)-based entailment checks, predominantly focus on surface-level similarity—lexical overlap, direct paraphrasing, or simple entailment. While effective for straightforward connections, these approaches falter when confronted with statements requiring multi-step inference or relying on implicit commonsense knowledge. This limitation becomes particularly problematic in scenarios requiring complex reasoning, like evaluating claim validity or analyzing argumentative discourse. In this paper, we propose a general reasoning-based framework that transcends simple entailment checking by examining the entire reasoning chain—both explicit and implicit steps—that connects statements. Our method maps reasoning paths for atomic statements, identifies unstated premises, and evaluates the validity of each premise to produce well-grounded judgments. Through comprehensive experiments on logical coherence, fallacy detection, and citation faithfulness benchmarks, we demonstrate that our approach provides superior performance compared to existing methods, offering a more granular and robust framework for assessing logical relationships between statements.

While Large Language Models (LLMs) excel at generating fluent text, they often struggle with multi-step logical reasoning with inferential connections among statements. We introduce KUDEX (Knowledge-Upward-Downward Reasoning Extension), a novel zero-shot, decompositional framework for analyzing and verifying logical relationships in natural language. Unlike traditional approaches that rely on surface-level entailment checks, KUDEX leverages LLMs to decompose reasoning into atomic primitives and constructs transparent reasoning graphs that serve both as execution traces and audit logs. Our framework includes a general prompt adaptation pipeline that adapts to diverse reasoning tasks without additional fine-tuning or domain-specific pipelines. Through comprehensive experiments on benchmarks across four domains, we demonstrate that KUDEX provides superior performance compared to existing methods, offering a more granular and robust framework for assessing logical relationships between statements.

\end{abstract}

\section{Introduction}

The advancement of Large Language Models (LLMs)~\cite{openai-chatgpt-online, anthropic2024claude, qin2024llm, touvron2023llama, deepseekai2024deepseekv3technicalreport} has revolutionized natural language understanding and generation, with systems now capable of producing lengthy, coherent text across diverse domains. 
%However, these capabilities bring new challenges in analyzing and verifying the logical relationships between statements. Currently, 
However, most methods for analyzing generated text rely on Natural Language Inference (NLI)-based entailment checks~\cite{devlin2019bert, liu2019roberta, he2020deberta, hu2024refchecker} that focus on surface-level similarity. While these approaches prove useful in certain contexts, they demonstrate significant limitations in scenarios requiring nuanced understanding of logical connections~\cite{wan2024logicasker, magomere2025finnli, sheng2025evaluating, patel2024multilogievalevaluatingmultisteplogical}, from academic discourse analysis~\cite{alhindi2023large} to argumentative reasoning~\cite{wadden2021multivers} and information verification~\cite{chakrabarty2021entrust}. 
%In these cases, users require more than surface-level assessment; they need confidence that the reasoning connecting statements is logically sound.

The principal challenge is that existing methods lack mechanisms to trace complex reasoning steps or identify missing assumptions that might silently bridge gaps in logic. When analyzing statements that require multiple inferential leaps to connect them, traditional entailment methods often fail to capture nuanced logical relationships. Even advanced reasoning-augmented approaches like Chain-of-Thought prompting \cite{wei2022chain} or commercial Large Reasoning Models (LRM)~\cite{anthropic2024claude} remain black-box methods, with latent chains lacking rigorous enforcement or external verification.
%This creates a significant blind spot in our ability to evaluate whether statements are truly connected through valid reasoning.

To address this gap, we propose KUDEX (Knowledge-Upward-Downward Reasoning Extension), a general reasoning-based framework that extends beyond simple entailment checking. Unlike existing approaches, KUDEX adopts a zero-shot, decompositional paradigm that leverages the power of LLMs. Instead of merely verifying whether statements match on a surface level, KUDEX evaluates the entire reasoning chain—both explicit and implicit steps—that connects them. By tagging reasoning units with atomic primitives (\textbf{K}nowledge, \textbf{U}pward generalization, and \textbf{D}ownward application) and connecting them as a structure, KUDEX assembles a fully transparent reasoning graph that serves both as the execution trace and audit log—enabling fine-grained verification and correction without additional fine-tuning or brittle domain-specific pipelines.

Our framework follows a principled two-pass annotation process: first labeling reasoning units with appropriate operation types and metadata, then constructing a reasoning graph that explicitly represents inferential dependencies. This structured approach makes reasoning paths transparent, shifting the focus from surface-level similarity to deeper reasoning integrity, which is applicable across diverse domains that require logical analysis.

%To address this gap, we propose a general reasoning-based framework that extends beyond simple entailment checking. Instead of merely verifying whether statements match on a surface level, our method evaluates the entire reasoning chain—both explicit and implicit steps—that connects them. By making these reasoning paths transparent, we shift the focus from surface-level similarity to deeper reasoning integrity, which is applicable across diverse domains requiring logical analysis.

Our contributions are as follows:

\begin{itemize}
    \item We introduce KUDEX, a zero-shot, decompositional reasoning paradigm that uses LLM solely as a parser to segment texts into atomic primitives, marks implied or delayed inferences, and assembles a fully transparent reasoning graph that enables fine-grained verification and correction without additional fine-tuning.
    
    %explicitly models and verifies reasoning chains between statements, including implicit premises and background knowledge through a minimal yet powerful set of reasoning primitives.
    \item We develop a general prompt adaptation pipeline that adapts KUDEX-Core to various reasoning tasks with much of the optimization and workload done by LLM.
    %\item We develop a two-stage algorithm that first maps reasoning paths for atomic statements and then assesses the verifiability of each premise to produce comprehensive judgments.
    \item We conduct extensive experiments across four different reasoning tasks: citation faithfulness evaluation, logical coherence assessment, logical reasoning in multiple-choice questions, and logical fallacy detection. By making reasoning paths transparent and verifiable,  KUDEX shows superior reasoning capability across all tasks compared to both LLM and LRM.
\end{itemize}
\vspace{-5pt}
\section{Related Works}
\vspace{-5pt}

Large language models (LLMs) such as GPT-4 \cite{openai2023gpt4}, Claude 3.7 \cite{anthropic2025claude37}, and DeepSeek-V3 \cite{deepseek2025v3} exhibit remarkable fluency and breadth in natural language generation but frequently fail at tasks requiring precise, multi-step reasoning.  Zero- or few-shot prompting often produces plausible but unsound justifications and outright hallucinations on arithmetic word problems \cite{cobbe2021training}, formal logic puzzles \cite{clark2020boolq}, and complex reading comprehension \cite{lin2022truthfulqa,ippolito2023controlled}.  To mitigate these shortcomings, “reasoning-augmented” pipelines such as Chain-of-Thought prompting \cite{wei2022chain}, ReAct \cite{yao2022react}, and Tree-of-Thoughts search \cite{yao2023tree} interleave explicit intermediate steps within generation, while commercial systems like Google Gemini \cite{google2024gemini} and Anthropic Claude \cite{anthropic2024claude} expose internal tool calls and reasoning modes. Despite improved accuracy on benchmarks such as GSM8K \cite{cobbe2021training} and HotpotQA \cite{yang2018hotpotqa}, these methods remain black-box: their latent chains lack rigorous enforcement or external verification and can propagate spurious inferences into final answers.

In parallel, neuro-symbolic and synthetic-corpus approaches aim to embed structured inference within or alongside LLMs.  Braid weaves fuzzy unification of symbolic predicates with a learned rule generator to produce interpretable proof graphs over commonsense stories (ROC Story Cloze) \cite{smith2021braid}.  Cora extracts frame-based causal graphs from unstructured biomedical and financial texts, then applies lightweight symbolic propagation and user-guided counterfactual interventions for high-stakes multi-hop inference \cite{lee2024multi}.  The Additional Logic Training (ALT) framework fine-tunes LLaMA-3 on a synthetic logic corpus (FLD×2), balancing positive/negative examples and diverse logical forms to yield up to +30 pt gains on formal logic benchmarks and notable improvements on math and NLI tasks \cite{zhao2024enhancing}.  Other efforts include Toolformer’s self-supervised API discovery \cite{wolf2023toolformer}, LLM-as-Program compilation of reasoning traces into executable code \cite{qin2024llm}, and LangChain’s orchestration of LLMs with external tools \cite{harrison2023langchain}.  In contrast, KUDEX is a zero-shot framework that leverages the language decomposition capabilities of LLM to build a fully transparent reasoning graph from distinct reasoning units.

%Unlike these, KUDEX-Core/PROOF adopts a zero-shot, decompositional paradigm: it uses an LLM solely as a parser to segment text into atomic \textbf{K}nowledge, \textbf{U}pward generalization, and \textbf{D}ownward application units, marks implied or delayed inferences, and assembles a fully transparent reasoning graph that serves both as the execution trace and audit log—enabling fine-grained verification and correction without additional fine-tuning or brittle domain-specific pipelines.

\vspace{-5pt}
\section{Method}

In this section, we first introduce the foundational KUDEX-Core framework upon which our methods are built. We then present a general prompt adaptation pipeline with Human-In-The-Loop (HITL) that extends KUDEX-Core to specific reasoning tasks zero-shot. Finally, we present an example of KUDEX adaptation, KUDEX-PF (proof), for the logical reasoning task.

%We propose a reasoning-based framework KUDEX for analyzing logical relationships between statements with LLMs. Unlike approaches that rely solely on surface-level similarity, our method examines the full reasoning chain, including both explicit and implicit steps, that connects the statements. In this section, we first introduce the foundational KUDEX-Core framework upon which our methods are built. We then present a general prompt adaptation pipeline with Human-In-The-Loop (HITL) that extends KUDEX-Core to specific reasoning tasks zero-shot. Finally, we present an example of KUDEX adaptation, KUDEX-PF (proof), for the logical reasoning task.
\vspace{-5pt}
\subsection{KUDEX-Core: The Foundation of Atomic Reasoning}
Natural language reasoning ranges from simple entailment checks to complex multi-hop inference chains, requiring logical verification at varying levels of granularity. Specialized approaches like Aristotelian logic~\cite{miller2015structure} and first-order predicate calculus~\cite{ertel2024first} struggle with the inherent fuzziness in natural languages due to their strict formal structures. To address this, we start from identifying elemental operations that form the basis of most logical relationships: stating foundational facts, extracting general principles from specific instances, and applying these principles to new contexts. 

\begin{figure}[H]
    \centering
    \includegraphics[width=\textwidth]{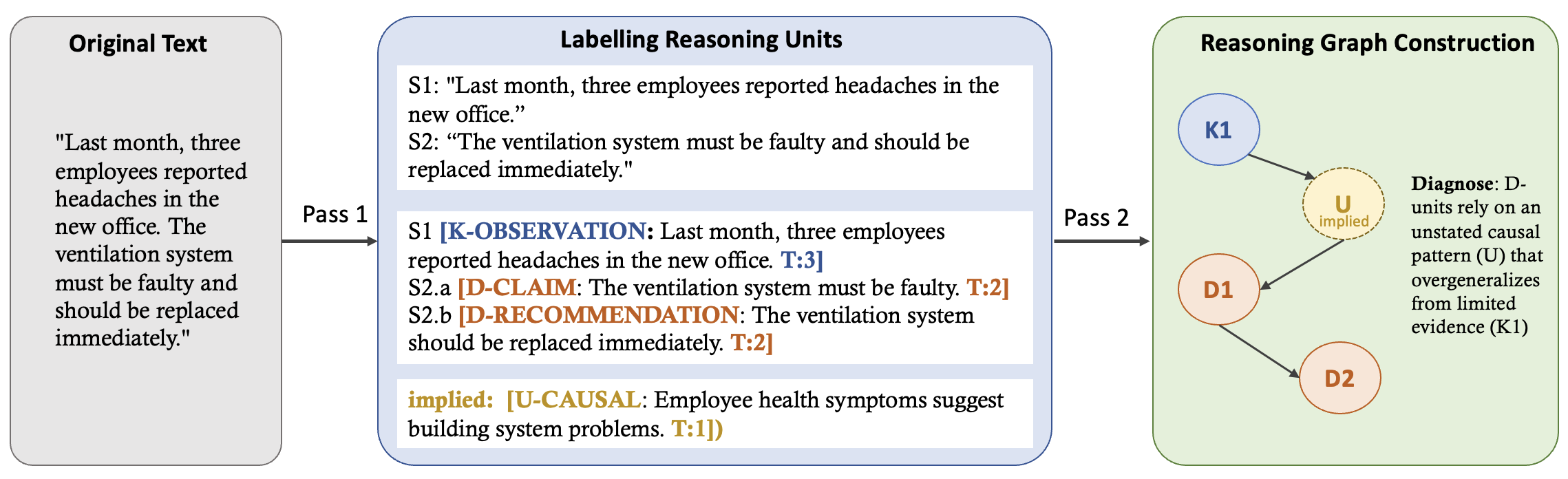}
    \caption{An example of two-pass annotation process of KUDEX-Core.}
    \label{fig:kudex_anno}
\end{figure}

\noindent \textbf{Reasoning Primitives.} The above observation leads us to propose three core primitives -- Knowledge (\textbf{K}), Upward generalization (\textbf{U}), and Downward application (\textbf{D}) -- that serve as the atomic units for constructing more complex reasoning structures.
As defined, (1) K represents factual statements, observations, evidence, or definitions that serve as the foundation for reasoning, capturing raw input information before any inferential processing occurs; (2) U infers a pattern or principle from one or multiple instances of K; (3) D applies generalizations to specific conclusions, producing claims, or interpretations. To illustrate these primitives, a basic inductive reasoning pattern can be expressed as: K (``Five batteries failed after two hours'' -- specific observation) $\rightarrow$ U (``Consistent early failure suggests design flaw'' -- general principle) $\rightarrow$ D (``The design is faulty'' -- specific conclusion). Figure~\ref{fig:kudex_anno} further presents some examples of how to annotate reasoning units with these primitives, with specific suffix (e.g., \texttt{K-ASSERTION} and \texttt{D-CLAIM}) indicating  the fine-grained label for each primitive.

%KUDEX-Core provides a minimal yet powerful set of reasoning primitives (\textbf{K-U-D}) that form the backbone of our approach. It decomposes complex reasoning into three fundamental operations: (1) \textbf{K(Knowledge)} represents factual statements, observations, evidence, or definitions that serve as the foundation for reasoning, capturing raw input information before any inferential processing occurs. (2) \textbf{U (Upward Generalization)}  represents inductive or abductive reasoning steps that extract patterns, principles, or generalizations from specific instances of \textbf{K}, moving from specific observations to broader principles. (3) \textbf{D (Downward Application)} represents deductive reasoning that applies generalizations to specific contexts, producing claims, warnings, recommendations, or interpretations, moving from general principles to specific conclusions. Figure~\ref{fig:kudex_anno} presents some examples of \textbf{K-U-D} primitives, and the suffix (e.g., \texttt{K-ASSERTION} and \texttt{D-CLAIM}) indicates the fine-grained label for each primitive, classified by LLM. 

For enhanced annotation precision, each primitive can be further qualified with two additional parameters. Attention (\textbf{A}) highlights key focus terms to indicate precisely what the reasoning is built from or applied to, marking the critical elements driving the inference. Temperature (\textbf{T}) indicates the level of certainty associated with a reasoning unit, ranging from highly speculative (\texttt{T:0}) to facts asserted with high confidence (\texttt{T:3}). Together, these primitives and qualifiers act as the foundation for decomposing and analyzing reasoning structures in KUDEX-Core.

\vspace{-2pt}
\noindent \textbf{Pass 1: Labeling the Reasoning Units.} In the first pass, KUDEX-Core enables decomposing complex reasoning into atomic units and annotates each following three steps: (1) Identify and Isolate Reasoning Units: Break down text into atomic reasoning acts, using hierarchical notation when multiple reasoning acts occur in a single statement (see S2.a, S2.b in Figure~\ref{fig:kudex_anno}). (2) Classify Operation Type and Add Descriptive Suffix: Determine whether each unit represents \textbf{K}, \textbf{U}, or \textbf{D}, and add a suffix describing its specific function (e.g., \texttt{K-EVIDENCE, U-PATTERN}). (3) Add Optional Metadata: Apply  attention markers \textbf{A} and temperature ratings \textbf{T} to indicate focus and certainty, respectively. (4) Add Supplementary Logic Marker:  Reasoning structures in natural language are often incomplete, rhetorically reordered, or unstated but necessary. KUDEX-Core also supports two resolution strategies to handle special types of \textbf{U}, as shown in Table~\ref{tab:special_u_cases}. These markers are not structural edges yet — they serve as hints for Pass 2.

\noindent \textbf{Pass 2: Reasoning Graph Construction.} In the second pass, we construct a reasoning graph by linking annotated units using directed edges that reflect inferential dependencies. This process enhances the traceability of reasoning and helps expose hidden gaps or circular logic.
%The result is a structured map of the reasoning flow — showing how claims are built, where support is strong or missing, and what overarching intent the passage pursues. 
Basically, the graph connects \textbf{K}, \textbf{U}, \textbf{D} nodes with directed edges that reflect their logical relationships. Natural language reasoning rarely follows a perfect linear path — arguments often contain unstated assumptions (acting as \textit{soft nodes}), out-of-order reasoning steps, or references to previous context. To capture these patterns, our graph structure includes two types of edges that mirror how humans actually reason.

The first type, \textit{inferential edges}, captures explicit reasoning steps present in the text. These include (1) \textbf{K}~$\rightarrow$~\textbf{U}, where factual knowledge supports a generalization (e.g., evidence leading to a pattern); (2) \textbf{U}~$\rightarrow$~\textbf{D}, where a generalization is applied to reach a specific conclusion;  and (3) \textbf{K}~$\rightarrow$~\textbf{D}, representing direct inference from facts to conclusions when intermediate generalizations are bypassed. 
The second type addresses \textit{implicit reasoning connections} that are essential for argument coherence but not explicitly articulated. When a claim logically require an unstated generalization, we create an \textbf{Implied U}~$\rightarrow$~\textbf{D} edge, connecting to a "soft node" representing the missing generalization. Similarly, when a conclusion appears before its supporting generalization in the text (a common rhetorical strategy), we use \textbf{Delayed U}$\rightarrow$\textbf{D} edges with forward links to maintain the logical structure.
Figure~\ref{fig:kudex_anno} further presents an example of applying KUDEX-Core analysis. 
%\tong{now the concepts, e.g. the two edge types and nodes being `soft', are a bit arbitrary, we can discuss about a better structure to motivate the design naturally.}

The two-pass annotation process of KUDEX-Core is performed by prompting an LLM with carefully designed instructions. For detailed prompt, please refer to Appendix~\ref{app:kudex-prompts}.
%\tong{we need to introduce that the above algorithm is implemented by prompting LLM, and give a pointer to the full prompt in the appendix.}
%\vspace{-40pt}
\begin{table}[H]
\centering
\caption{Handling Special U-Type Cases}
\label{tab:special_u_cases}
\begin{tabular}{p{0.15\linewidth}p{0.35\linewidth}p{0.4\linewidth}}
\toprule
\textbf{Type} & \textbf{What It Means} & \textbf{How to Mark It} \\
\midrule
\textbf{Implied U} & A D-unit depends on a generalization that is \textbf{not stated} but clearly required. & Add a parenthetical marker: \texttt{(implied: [U-...])} \\
\midrule
\textbf{Delayed U} & The U (or K) a D-unit depends on appears \textbf{later} in the text. & Add a marker: \texttt{(delayed-U): \text{See St}}, where St is the sentence appeas later\\
\bottomrule
\end{tabular}
\end{table}
%\vspace{-20pt}
\subsection{Prompt Adaptation Pipeline}
\begin{figure}[H]
    \centering
    \includegraphics[width=0.85\textwidth]{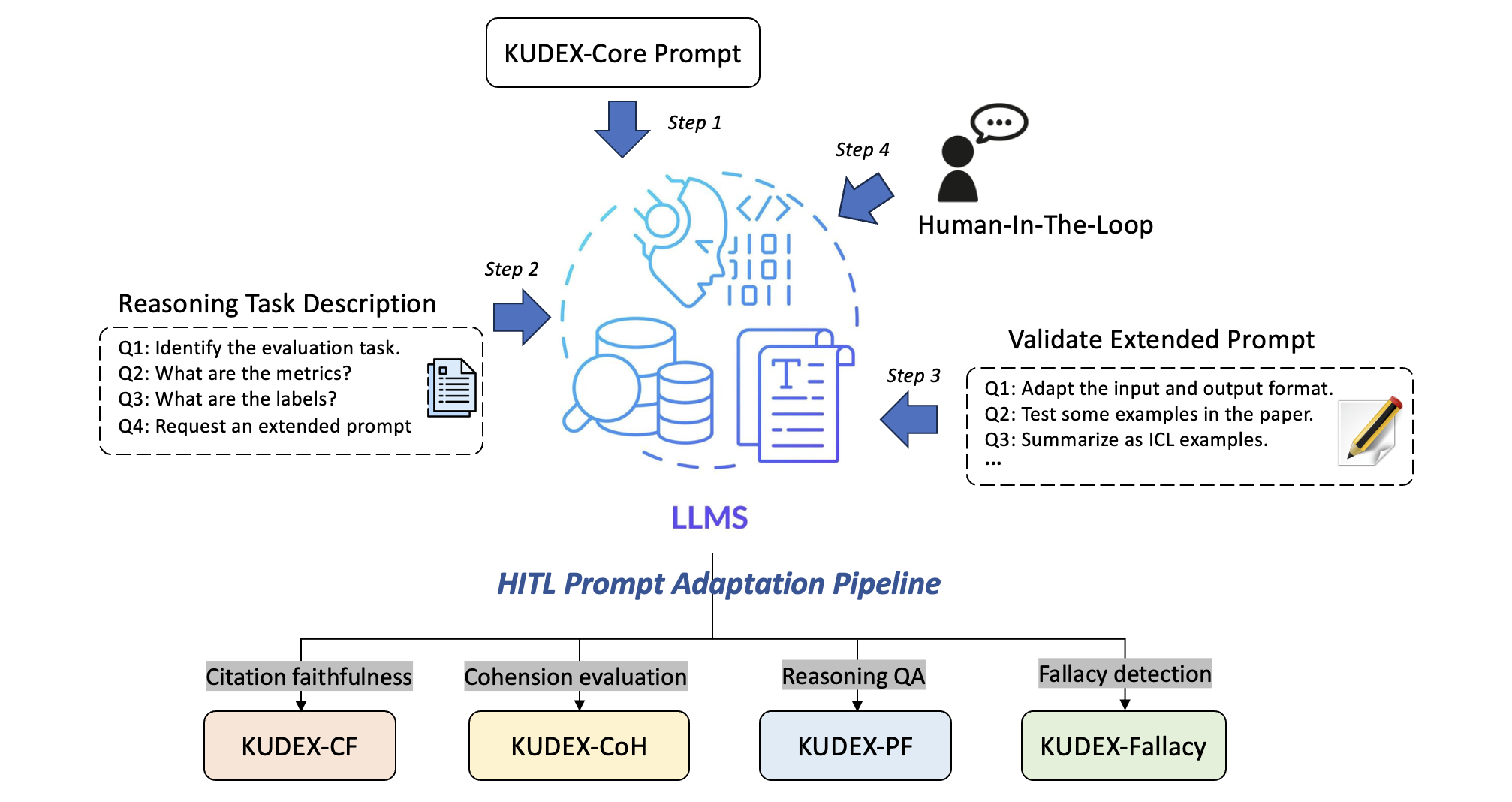}
    \caption{An overview of the prompt adaptation pipeline with HITL, which extends KUDEX-Core to various reasoning tasks, e.g., reasoning QA~\cite{liu2020logiqa}, cohesion evaluation~\cite{maimon2023cohesentia}, etc.}
    \label{fig:adapt_prompt}
\end{figure}

%\vspace{-15pt}
\label{sec: prompt_ada}
KUDEX-Core provides a foundation for textual decomposition and logical annotation; however, adaptation is required when applying it to specific reasoning tasks with distinct logical operations and annotation requirements. To address this need, we propose a Large Language Model (LLM)-based Human-In-The-Loop (HITL) pipeline comprising four key steps to extend to a new prompt, as illustrated in Figure~\ref{fig:adapt_prompt}. First, we initialize the process by presenting the KUDEX-Core prompt to an LLM to establish the baseline framework. Second, we introduce the specific reasoning task description to the LLM and request an extended prompt tailored to the task requirements. Third, we validate the extended prompt by applying it to representative sample data from the target task. Finally, based on the results, a human expert evaluates the result and determines whether to proceed with the current prompt or iterate from the second step. This pipeline yields a task-specific LLM prompt that effectively extends KUDEX-Core's capabilities to address the particular problem domain.

%(1) digest the KUDEX framework, (2) read an extension, typically a paper associated with a specific benchmark, (3) warm up by listing a few illustrative examples, and (4) get ready to take and data point and make judgment — all achieved through interaction with a large language model (LLM). \tong{we should introduce the general purpose of each step here, rather than the exact technique (e.g. have it read a paper or some other materials).} Figure~\ref{fig:adapt_prompt} illustrates how to flexibly adapt KUDEX-Core prompt to various reasoning tasks, with much of the optimization and workload offloaded to the LLM. 

\subsection{KUDEX-PF: Extension for Formal Reasoning Tasks}
Following the prompt adaptation pipeline in Section~\ref{sec: prompt_ada}, we present one extension, KUDEX-PF, a proof version of KUDEX-Core for formal logical reasoning task. This task requires complex multi-step reasoning towards the answer and thus the adaptation is needed to formulate a more structured logical inference with identifiable logical operations. To achieve this, we first let LLM to digest the KUDEX-Core prompt, and then extend the prompt by learning the new task from a benchmark paper, e.g., LogiQA 2.0~\cite{, liu2023logiqa}. Then we validate the extended prompt by a quick test on several illustrative data samples extracted from the paper. We finally get the KUDEX-PF prompt with human examination along the process.

\textbf{Key Extensions.} First, we add a new primitive Reasoning Step (\textbf{R}), capturing explicit inference steps that connect other units via identifiable logical operations. Second, \textbf{formal inference rule tags} have been incoporated to identify standard patterns of deductive logic, e.g., \texttt{[MP]} for Modus Ponens, \texttt{[MT]} for Modus Tollens, \texttt{[UI]} for Universal Instantiation,  \texttt{[HS]} for Hypothetical Syllogism, etc. Third, a set of \textbf{structural tags} are introduced to better organize multi-step reasoning, e.g.,\texttt{U-intermediate} marks an intermediate inference required for chaining; \texttt{D-final} indicates the final conclusion of an argument; and \texttt{proof-type} specifies the direction of reasoning, such as forward or backward chaining.
\vspace{-10pt}

\noindent \textbf{Two-Pass Annotation Process.} KUDEX-PF follows a two-pass annotation process, first applying KUDEX-Core analysis and then adding \textbf{R} nodes and formal reasoning extensions. In the second pass, \textbf{R} nodes are introduced to represent explicit inference steps that link existing \textbf{K}, \textbf{U}, and \textbf{D} units. After tagging them with 
appropriate formal inference rule labels, structural tags are also added to mark the presence of logical constructs like quantifiers, variables, and proof structure. Finally, this pass involves identifying more complex reasoning patterns, including non-monotonic inferences, where conclusions might be retracted in light of new information. Table~\ref{tab:kudex_pf_example} in Appendix~\ref{app:method} shows an example of the annotation results using KUDEX-PF.

KUDEX-PF explicitly captures the logical structure of deductive arguments, making it particularly well-suited for tasks requiring step-by-step logical verification. Its structural representation enables rigorous assessment of proof validity beyond KUDEX-Core.
\vspace{-10pt}
\section{Experiments}
\vspace{-5pt}
We extensively evaluate our general reasoning framework on four different reasoning tasks, including: (1) \textit{citation faithfulness evaluation} in AI-generated reports, (2) \textit{logical coherence evaluation} in AI generated stories, (3) \textit{logical reasoning} in logical reasoning
reading comprehension tests, and (4) \textit{logical fallacy detection} in sentences with fallacy statements.

\vspace{-5pt}
\subsection{Datasets}
\vspace{-3pt}

\noindent \textbf{Citation Faithfulness Bench} is an in-house dataset designed to evaluate citation faithfulness in AI-generated reports. Curated by AWS Q Business, it is annotated by two annotators for 130 claim-source pairs, each labeled as  \textit{Fully Support, Partially Support, Neutral,} or \textit{Contradict}. The claim-source pairs are drawn from a wide range of reports generated by OpenAI Deep Research~\cite{openai2024deepresearch}, Perplexity~\cite{perplexity2024}, and Deep Thought, a system developed by AWS Q Business, covering topics such as finance, sports, technology, management, and more. Unlike traditional Natural Language Inference (NLI) tasks, this evaluation task demands nuanced reasoning to align each claim with its source, interpret implicit premises, find evidence conflicts and account for domain-specific context and terminology. The benchmark will be open-sourced following the launch of Deep Thought.

\noindent \textbf{CoheSentia}~\cite{maimon2023cohesentia} is a benchmark designed to assess logical coherence in AI-generated stories. The dataset comprises 342 sentences, each paired with its preceding paragraph and annotated for the coherence issue between the two parts. For incoherent sentences, the benchmark identifies specific issues across three dimensions: cohesion (text flow), consistency (logical alignment), and relevance (topical appropriateness).

\noindent \textbf{LogiQA 2.0}~\cite{liu2023logiqa} is a challenging logical reasoning benchmark in natural language inference, with a larger scale of data and fine-grained annotations compared with LogiQA~\cite{liu2020logiqa} released in 2020. The test set consists of 3240 premise-hypothesis pairs translated from Chinese Civil Service Examinations, each pair demands sophisticated multi‐sentence logical inference, keen analytical reasoning, and explicit consideration of implicit assumptions.

\noindent \textbf{SmartyPat-Bench}~\cite{xu2025socrates} is a comprehensive and recently released benchmark for logical fallacy detection, comprising 502 high-quality, naturally expressed English sentences containing subtle logical fallacies, manually curated from real-world Reddit posts and systematically labeled with one or more of 14 fallacy types. To address fallacy-type imbalance and scale beyond manual collection, the same work introduces SmartyPat-Bench-Augmented, an expanded dataset automatically synthesized by the SmartyPat framework using Prolog-based oracles to generate additional fallacious statements, ensuring balanced coverage across all fallacy categories. We employ both datasets in our experiments.

\vspace{-5pt}
\subsection{Reasoning Capability Across Benchmarks}
\vspace{-3pt}

We compare Claude 3.7-based KUDEX against two other baselines: (1) LLM: standard prompting with Claude 3.7 Sonnet~\cite{anthropic2024claude}, and (2) LRM: Claude 3.7 Sonnet with extended thinking. The key distinction is that KUDEX generates task-specific adaptive prompts, creating specialized variants (KUDEX-CF, KUDEX-PF, KUDEX-CoH, and KUDEX-Fallacy), whereas the baselines simply use modified prompts tailored to each target task. For more details of prompts and complete results, please refer to Appendix \ref{app:prompts} and \ref{app: detailed_results}.

\begin{wraptable}{r}{0.4\textwidth}
\centering
\caption{Citation Faithfulness Results}
\label{tab:citation_faithfulness_results}
\begin{tabular}{lc}
\toprule
\textbf{Method} & \textbf{Accuracy} \\
\midrule
LLM & 0.7462 \\
LRM & 0.7615 \\
KUDEX-CF & \textbf{0.8268} \\
\bottomrule
\end{tabular}
\end{wraptable}

\noindent \textbf{Citation Faithfulness.} Table~\ref{tab:citation_faithfulness_results} presents the citation faithfulness accuracy, where KUDEX-CF significantly outperforms both baselines. The substantial performance gain demonstrates the effectiveness of KUDEX-CF in evaluating citation faithfulness in AI-generated reports. KUDEX-Core's decomposition framework provides particular advantages for citation evaluation by clearly separating source knowledge (K) from inferential generalizations (U) and specific applications (D), thus to trace how evidence connects to the evaluated statement.

\noindent \textbf{Logical Coherence.} Table~\ref{tab:cohesentia_results} shows that KUDEX-CoH achieves the highest F1 scores across most dimensions—with significant improvements in overall coherence and two sub-dimensions: cohesion and consistency. While the baselines show slightly higher F1 score in relevance detection, further analysis (Appendix~\ref{app:kudex-relevance}) suggests this stems not from a shortcoming of KUDEX, but from its stricter enforcement of topical alignment. KUDEX flags subtle relevance violations that are often overlooked by human annotators and simpler MBE methods.

\begin{table*}[h!]
\centering
\caption{F1 Score comparison of different reasoning methods on Cohesentia benchmark.}
\label{tab:cohesentia_results}
\begin{threeparttable}
\begin{tabular}{lccc}
\toprule
\textbf{Dimension} & \textbf{LLM} & \textbf{LRM} & \textbf{KUDEX-CoH} \\
\midrule
Coherence & 0.6615 & 0.6719 & \textbf{0.7167} \\
\midrule
Cohesion & 0.4889 & 0.5474 & \textbf{0.5600} \\
Consistency & 0.4634 & 0.5195 & \textbf{0.5432} \\
Relevance & 0.4444 & \textbf{0.5000} & 0.4000 \\
\bottomrule
\end{tabular}%
\end{threeparttable}
\end{table*}
\vspace{-3pt}

\noindent \textbf{Logical Reasoning.} As shown in Table~\ref{tab:logiqa_results}, KUDEX-PF achieves the highest accuracy and F1 score. These results demonstrate that our structured approach to logical inference, with explicit reasoning steps and formal inference rule tagging, significantly enhances performance on complex reasoning tasks that require precise deductive chains.
%The results validate our hypothesis that decomposing complex logical problems into explicitly tagged inference steps helps the model navigate through challenging multi-step reasoning problems more effectively than traditional prompting methods.
\vspace{-3pt}

\begin{table}[h!]
\centering
\caption{Accuracy and F1 Score comparison of different reasoning methods on LogiQA2.0 dataset.}
\label{tab:logiqa_results}
\begin{threeparttable}
\begin{tabular}{lccc}
\toprule
\textbf{Metric} & \textbf{LLM} & \textbf{LRM} & \textbf{KUDEX-PF} \\
\midrule
Accuracy  & 0.6820 & 0.6770 & \textbf{0.7010} \\
%\hline
F1 Score  & 0.6750 & 0.6620 & \textbf{0.7010} \\
\bottomrule
\end{tabular}
\end{threeparttable}
\end{table}
\vspace{-3pt}

\noindent \textbf{Logical Fallacy Detection.} As shown in Tables~\ref{tab:fallacy_results}, KUDEX-Fallacy consistently outperforms baseline methods across both evaluation datasets. KUDEX-Fallacy maintains superior F1 scores over both the original SmartyPat-Bench and balanced SmartyPatBench-Augmented. These advantages demonstrate the effectiveness of breaking down each fallacy type in the reasoning pipeline. For example: \textit{false analogies} represent flawed K-to-U generalizations, 
%\textit{equivocation} indicates inconsistent K definitions across reasoning steps, 
and \textit{inverse errors} involve improper D applications from generalizations. By systematically analyzing how each primitive interacts within an argument, KUDEX-Fallacy more precisely identifies logical errors that traditional approaches often miss.

\vspace{-3pt}
\begin{table}[h!]
\centering
\caption{F1 Score comparison across different reasoning methods on fallacy detection datasets.}
\label{tab:fallacy_results}
\begin{threeparttable}
\begin{tabular}{lcccccc}
\toprule
\multirow{2}{*}{\textbf{Metric}} & \multicolumn{3}{c}{\textbf{SmartyPat-Bench}} & \multicolumn{3}{c}{\textbf{SmartyPatBench-Augmented}} \\
\cmidrule(r){2-4}\cmidrule(r){5-7}
 & \textbf{LLM} & \textbf{LRM} & \textbf{KUDEX-Fallacy} & \textbf{LLM} & \textbf{LRM} & \textbf{KUDEX-Fallacy} \\
 \cmidrule(r){1-1} \cmidrule(r){2-4} \cmidrule(r){5-7}
%\hline
Macro avg F1 & 0.3265 & 0.2624 & \textbf{0.3683} & \multirow{2}{*}{0.6248} & \multirow{2}{*}{0.5529} & \multirow{2}{*}{\textbf{0.6524}} \\
%\cline{1-4}
Weighted avg F1 & 0.5645 & 0.5204 & \textbf{0.5799} & & & \\
\bottomrule
\end{tabular}
\end{threeparttable}
\end{table}
\vspace{-5pt}

\subsection{In-Depth Analysis}
\noindent \textbf{Structured Reasoning vs. Built-in Reasoning.} 
%Our experimental results reveal that a standard LLM equipped with structured, post-hoc reasoning framework (KUDEX) consistently outperforms LLMs with built-in reasoning modes (LRM).
The performance gap between our structured reasoning framework and built-in reasoning baselines can be attributed to several factors. First, explicit reasoning decomposition provides better traceability and verification, while black-box reasoning may internally propagate errors without the opportunity for correction. Second, our reasoning graph construction allows for detection of hidden assumptions and circular logic that might go unnoticed in monolithic reasoning approaches. Third, decomposed primitive reasoning units enable more precise identification of failures, whereas built-in reasoning modes typically provide only the final result without exposing the underlying inference structure.

\noindent \textbf{Adaptation Effort Across Tasks.} While KUDEX offers zero-shot adaptability across different reasoning tasks, the HITL few-shot effort required varies by domain. KUDEX-PF for logical reasoning required minimal adaptation effort, as formal logical structures align naturally with our K-U-D primitives and inference rules. In contrast, KUDEX-Fallacy required more extensive prompt engineering to handle the diverse manifestations of fallacy detection. KUDEX-CF for citation faithfulness fell in the middle, requiring moderate customization to handle the specific relationship patterns between sources and claims. 
This variation highlights a trade-off between framework generality and task-specific optimization, though the zero-shot nature of KUDEX significantly reduces the overall effort compared to supervised approaches.

\begin{figure}[htbp]
    \centering
    \includegraphics[width=0.8\textwidth]{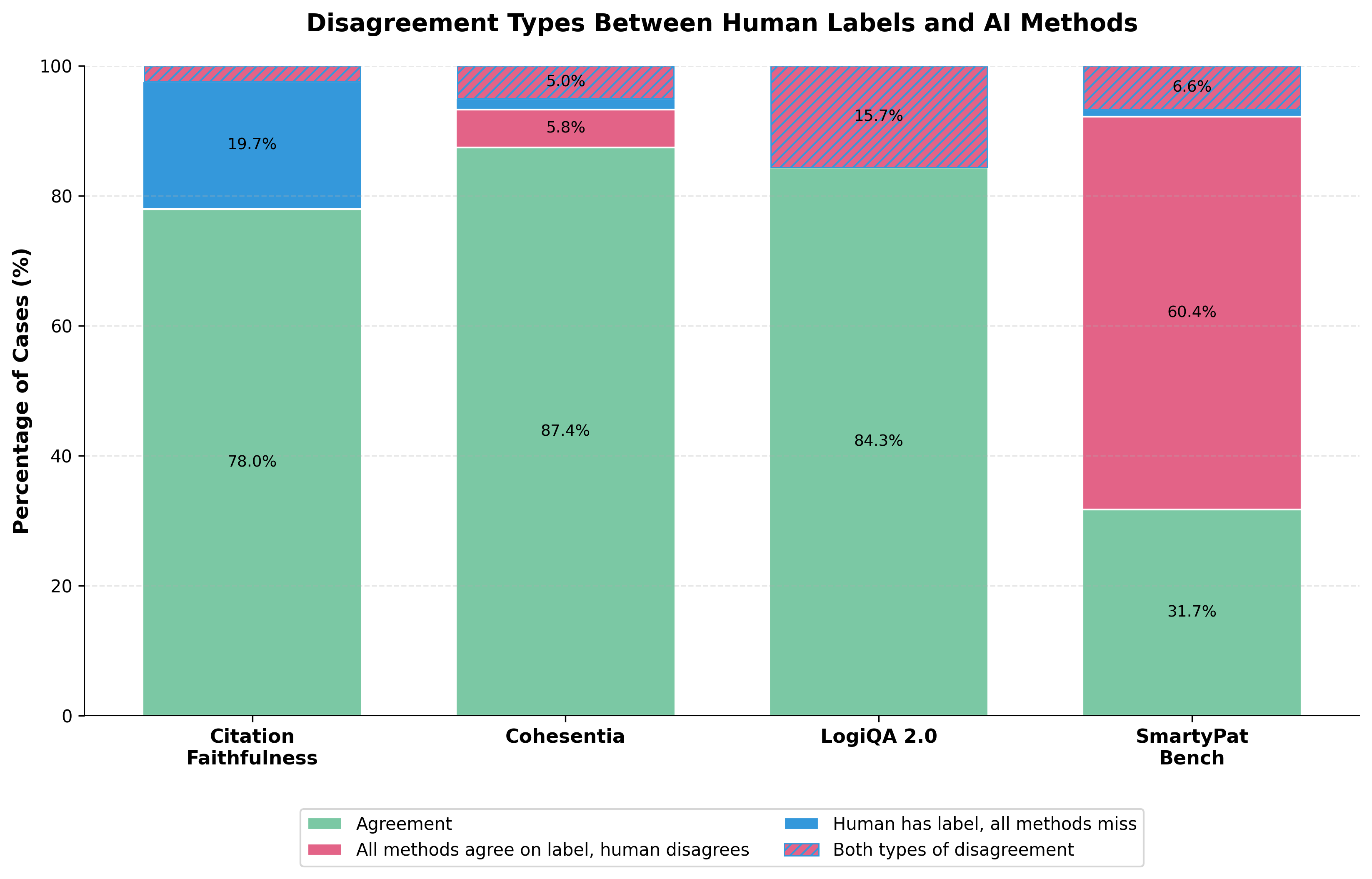}
    \caption{Disagreement patterns between human annotations and different methods across four benchmarks. The bars show the distribution of labeling agreement and disagreements: \textcolor{typegreen}{agreement cases}, \textcolor{typeblue}{instances where humans assign labels that all methods miss}, \textcolor{typepink}{situations where all methods converge on labels not assigned by humans} and \textcolor{typepurple}{cases where both disagreement types occur simultaneously} (stripped). For LogiQA 2.0's binary classification, all disagreements fall into the overlapping category. While for multi-label tasks (Cohesentia and SmartyPat-Bench), cases where all three models converge on labels not assigned by humans outnumber those where humans assign labels missed by all models, suggesting that human annotators often overlook valid categories. This pattern is less pronounced in the Citation Faithfulness dataset, which involves single-label prediction.}
    \label{fig:disagreement}
\end{figure}

\noindent \textbf{Human Labels Are Not Always Trustworthy.} We observed significant disagreements between MBE methods and human annotations as shown in Figure~\ref{fig:disagreement}, which aligns with recent research showing that LLMs can outperform humans on complex reasoning and factuality tasks~\cite{wei2024longformfactualitylargelanguage}. Specifically, in 15.7\% of LogiQA cases and 60.4\% of SmartyPat-Bench instances, all three model variants (LLM, LRM, and KUDEX) unanimously disagreed with the original human labels. Manual inspection (Appendix~\ref{app:case-study}) found that model predictions were frequently more accurate—revealing overlooked constraints, richer inference paths, or logical flaws missed by annotators. These findings suggest that many observed disagreements arise from annotation gaps rather than model error. They also highlight the potential of model-based reasoning not only for solving tasks, but for auditing and improving benchmark quality. Enhancing annotation protocols, e.g., by supporting multi-label justifications, extending annotation time, or incorporating model-in-the-loop verification, could significantly reduce such discrepancies.

\vspace{-15pt}
\section{Conclusion}
\vspace{-5pt}
In this work, we introduced KUDEX, a modular reasoning framework that decomposes complex inference into explicit Knowledge–Understanding–Deduction primitives, enabling transparent chain-of-thought tracing and task-specific adaptation. Across four diverse benchmarks—including citation faithfulness, logical coherence, multi-step inference, and fallacy detection—KUDEX consistently outperformed both standard and extended-thinking baselines. Our in-depth analysis highlights the trade-offs between zero-shot generality and prompt customization effort, as well as the potential of model-driven audits to uncover annotation inconsistencies. Future work will explore tighter integration with formal verification tools, automated prompt optimization, and broader evaluations on real-world NLP deployments to further enhance the reliability and interpretability of language model reasoning.
\vspace{-5pt}
\section{Customer Problem Statement}
\vspace{-5pt}
While Large Language Models (LLMs) have shown remarkable capabilities across various applications, ensuring reliable reasoning remains a critical challenge that impacts customer trust and application safety. Organizations deploying LLM-based solutions need robust reasoning frameworks to safeguard the correctness and credibility of their applications. The KUDEX framework addresses this fundamental need by enhancing LLM reasoning capabilities, benefiting customers across different use cases. However, like any reasoning system, KUDEX is not infallible and can occasionally produce incorrect inferences, highlighting the need for continuous improvement. To address these limitations, the framework can be enhanced through model fine-tuning to internalize reasoning patterns and integration with complementary neural-symbolic techniques, ultimately working toward more reliable and trustworthy AI systems.

\bibliography{citation}
\bibliographystyle{ieee_citation_format}

\appendix

\newpage
\section*{Appendix}

\section{Method Explanation}
\label{app:method}

\begin{table}[H]
\centering
\caption{KUDEX-Core Primitives}
\label{tab:kudex_core}
\begin{tabular}{p{0.15\linewidth}p{0.25\linewidth}p{0.22\linewidth}p{0.25\linewidth}}
\toprule
\textbf{Operation} & \textbf{Definition} & \textbf{Suffix} & \textbf{Example} \\
\midrule
\textbf{K}: Knowledge & States a fact, belief, perception, definition, or intention & \texttt{K-EVIDENCE}, \texttt{K-ASSERTION, K-DEFINITION, K-INTENTION} & [\texttt{K-EVIDENCE}:"Five batteries failed after two hours."] \\
\midrule
\textbf{U}: Upward Generalization & Infers a pattern or principle from multiple instances of K, and includes causal, temporal, conditional, analogical, or statistical reasoning. & \texttt{U-PATTERN, U-CAUSAL, U-TEMPORAL, U-ANALOGY} & [\texttt{U-PATTERN}:"Consistent early failure suggests design flaw."] \\
\midrule
\textbf{D}: Downward Application & Applies a generalization to formulate a claim, warning, prediction, or interpretation & \texttt{D-CLAIM, D-WARNING, D-RECOMMENDATION, D-IMPLICATION} & [\texttt{D-CLAIM}:"The design is faulty."] \\
\bottomrule
\end{tabular}
\end{table}

\begin{table}[h]
\centering
\caption{Example Application of KUDEX-PF Analysis}
\label{tab:kudex_pf_example}
\begin{tabular}{p{0.2\linewidth}p{0.7\linewidth}}
\toprule
\textbf{Original Text} & "All birds have wings. Eagles are birds. Therefore, eagles have wings." \\
\midrule
\textbf{Annotation} & \texttt{K1 [PREMISE]}: All birds have wings. \texttt{[T:3]} \\
& \texttt{K2 [PREMISE]}: Eagles are birds. \texttt{[T:3]} \\
& \\
& \hspace{1em} \texttt{R1 [UI]}: From K1, instantiate universal statement to "If Eagle is a bird, then Eagle has wings." \\
& \hspace{2em} • \texttt{U1 [INTERMEDIATE]}: bird(Eagle) → wings(Eagle) \texttt{[T:3]} \\
& \\
& \hspace{1em} \texttt{R2 [MP]}: From U1 and K2, apply Modus Ponens. \\
& \hspace{2em} • \texttt{D1 [FINAL]}: Therefore, eagles have wings. [\texttt{T:3]} \\
\bottomrule
\end{tabular}
\end{table}

\section{Dive Deep Into Disagreement Between Human Annotators and \textsc{MBE}}
\label{app:case-study}

We manually reviewed two recurring kinds of mismatches in every benchmark—situations in which (i) all three systems converge on the same answer while the gold label diverges, and (ii) the annotator chooses an answer that none of the MBE methods propose. Representative examples are presented below to diagnose the roots of these discrepancies and to illuminate the contrasting reasoning habits of humans and large language models.

%------------------------------------------------------------  
\subsection{Citation~Faithfulness}

\vspace{2pt}
\noindent\textbf{Claim.}  
\emph{``Organizations should evaluate pricing structures carefully.''}

\noindent\textbf{Human label:} \textsc{Neutral}\hfill \\
\textbf{All three systems:} \textsc{Supported}

\paragraph{Evidence snippets (condensed).}
\begin{itemize}
  \item {\small\emph{``Both Google Apigee and Azure API Management offer flexible, tiered pricing models tailored to different business needs.''}}
  \item {\small\emph{``Pricing and licensing models … plans range from evaluation to enterprise tiers based on call volume and support level.''}}
  \item {\small\emph{``FAQ: How do they compare in terms of pricing?'' \;and\; ``Pay-as-you-go options ensure that businesses only pay for what they use.''}}
\end{itemize}

\paragraph{Explanation.}
The source document is a lengthy article, and it dose not state the claim explicitly.  
A human annotator, likely scanning for direct matches, may overlook relevant evidence that requires inference rather than surface-level alignment—ultimately labeling the claim as \textsc{Neutral}.  
In contrast, LLMs successfully synthesize clues from:

\begin{itemize}
  \item[(a)] a section that discusses multiple pricing tiers for different business needs;
  \item[(b)] remarks about pay-as-you-go models tied to usage patterns; and
  \item[(c)] an FAQ entry emphasizing price comparison as a key concern.
\end{itemize}

By combining these signals, the models reasonably infer that careful evaluation of pricing structures is warranted.  
This case illustrates how citation faithfulness often requires multi-hop, pragmatic reasoning—something human annotators may miss under time constraints, but which LLMs can handle effectively.

%------------------------------------------------------------  
\subsection{Logical~Coherence}

\begin{enumerate}[label=\textbf{C\arabic*}.]

\item \textbf{Anachronism}

\textbf{Input.}
\begin{quote}\small
\textbf{Title:} \emph{The War for Humanity: The War in Iraq}\\
\textbf{Paragraph:} It was the early morning hours of March 19, 2020.\\
\textbf{Sentence:} The sounds of American and British war planes could be heard overhead
as they began their bombing campaign against the Iraqi capital of Baghdad.
\end{quote}

\textbf{Human label:} \textsc{no\_error}\quad \\
\textbf{All three systems:} \textsc{coherence\_error}

\textbf{Explanation:}
Coherence here hinges on world knowledge: the U.S.–U.K. bombing of Baghdad began
in \textbf{2003}, not 2020.  
All three models detect this temporal clash and flag a world knowledge inconsistency error.
Human annotators, focused on local narrative flow, either overlooked the date or lacked
the historical fact—illustrating how external knowledge is essential for this
benchmark.

\item \textbf{Gender swap}

\textbf{Input.}
\begin{quote}\small
\textbf{Title:} \emph{The Witch of the Dark Woods}\\
\textbf{Paragraph:} Deep in the forest there lives a Witch. She is a cruel woman who takes
pleasure in torturing animals and children. If you anger her, she will curse you with her magic.
The villagers live in fear of her and do their best to avoid her wrath.\\
\textbf{Sentence:} One day, a young \textbf{girl} named Hansel and \textbf{his} sister Gretel
stumble into the Witch's domain while they are playing hide-and-seek.
\end{quote}

\textbf{Human label:} \textsc{no\_error}\quad \\
\textbf{All three systems:} \textsc{coherence\_error}

\textbf{Explanation:}
The sentence violates gender consistency: Hansel—canonically male—is
called a \emph{girl}, and the pronoun \emph{his} is used immediately afterward.
LLMs catch both the pronoun–noun mismatch and the clash with the well-known
fairy-tale schema.  
Human annotators, reading quickly, often overlook such micro-level
inconsistencies that require fine-grained coreference resolution.

\item \textbf{Over-penalised summary}

\textbf{Input.}
\begin{quote}\small
\textbf{Title:} \emph{A Wonderful Day}\\
\textbf{Paragraph:} There were two wonderful people in that day. One was the
salesperson at the store, and the other was a grandma who walked her grandson
to school.\\
\textbf{Sentence:} That's all they did for most of the day.
\end{quote}

\textbf{Human label:} \textsc{relevance\_error}\quad \\
\textbf{All three systems:} \textsc{no\_error}

\textbf{Explanation:}
The sentence succinctly summarises the preceding paragraph and is therefore
fully relevant.  
The annotator likely mis-interpreted the deictic \emph{they} or treated summary
sentences as filler, wrongly tagging it as off-topic.  
The models, which track discourse structure, judge the sentence acceptable.

\end{enumerate}

%------------------------------------------------------------  
\subsection{Logical~Reasoning}

\begin{enumerate}[label=\textbf{R\arabic*}.]

\item \textbf{Underdetermined hiring puzzle}

\textbf{Input.}
\begin{quote}\small
\textbf{Premise:}\\
Five graduate students (Zhang, Wang, Li, Zhao, Liu) were hired by one of three
employers (Tianshu, Tianji, Tianxuan), and each employer hired at least one
student.\\
(1) Li was hired by Tianshu; \quad
(2) Li and Zhao were not employed by the same unit;\\
(3) Liu and Zhao were hired by the same unit; \quad
(4) If Zhang was hired by Tianxuan, Wang was also hired by Tianxuan.\\[2pt]
\textbf{Hypothesis:}\\
“Zhang was hired by Tianxuan” is correct to determine the employer of each
graduate.
\end{quote}

\textbf{Human label:} \textsc{entailment}\quad \\
\textbf{All three systems:} \textsc{not-entailment}

\begin{table}[h!]
\centering
\small
\begin{tabular}{lcc}
\toprule
\textbf{Tianshu} & \textbf{Tianji} & \textbf{Tianxuan}\\
\midrule
Li & Zhao,\,Liu & Zhang,\,Wang\\
Li,\,Zhang & Zhao,\,Liu & Wang\\
Li & Zhang & Zhao,\,Liu,\,Wang\\
\bottomrule
\end{tabular}
\caption{Three concrete hiring patterns: Zhang can work in \emph{any} of the three companies without violating the four constraints.}
\label{tab:hiring-patterns}
\end{table}

\textbf{Explanation:}
The annotator concluded the hypothesis is correct, but exhaustive search shows at
least three valid assignments as shown in Table~\ref{tab:hiring-patterns} that place Zhang
in \emph{each} of the three companies.  
Because multiple solutions exist, the statement is \textsc{not-entailed}.  
Humans often anchor on the first plausible arrangement and stop—a classic
satisficing error—whereas LLMs can systematically enumerate the full solution
space.

\item \textbf{Hip-hop on the square}

\textbf{Input.}
\begin{quote}\small
\textbf{Premise:}\\
Collective behavior is a spontaneous and unorganized behavior with a large
number of people. In collective behavior, individuals do not act independently,
but rely on and influence each other.\\[2pt]
\textbf{Hypothesis:}\\
According to the above definition, dancing hip-hop in the open space in the
evening is \emph{not} a collective behavior.
\end{quote}

\textbf{Human label:} \textsc{entailment}\quad \\
\textbf{All three systems:} \textsc{not-entailment}

\textbf{Explanation:}
The premise merely \emph{defines} collective behaviour; it provides no facts
about whether the evening hip-hop session is spontaneous, organised, large, or
interactive.  
Models correctly note that the definition alone cannot justify the hypothesis,
whereas the annotator imported unstated commonsense assumptions.

\item \textbf{Cruise-ship width vs.\ cabin size}

\textbf{Input.}
\begin{quote}\small
\textbf{Premise:}\\
British cruise ships were initially unpopular; French ships sold better because
they had more open space.  
Analysis showed passengers care more about public-space width than cabin size,
so the British shrank cabins and widened public areas, leading to strong sales.\\[2pt]
\textbf{Hypothesis:}\\
“The moderation of width and narrowness is what we really need” is an inaccurate
understanding of this passage.
\end{quote}

\textbf{Human label:} \textsc{not-entailment}\quad \\
\textbf{All three systems:} \textsc{entailment}

\textbf{Explanation:}
The passage endorses a \emph{priority}, not a “moderation,” between cabin width
and public-space width: widen common areas, shrink rooms, sales improve.
Therefore the hypothesis—stating that “moderation” misreads the text—is indeed
supported, and the models label \textsc{entailment}.  
The annotator likely fixated on the absent word “moderation” and the implied prioritization.

\end{enumerate}

%------------------------------------------------------------  
\subsection{Fallacy~Detection}

\begin{enumerate}[label=\textbf{F\arabic*}.]

\item \textbf{“Smart TV” argument}

\textbf{Input.}
\begin{quote}\small
Since a smart TV has intelligence like human beings, and “smart” means better
computational power than a traditional TV, and \emph{Duck Dynasty} is considered
a less intelligent show, and the TV can still watch \emph{Duck Dynasty}, then
the TV is not truly smart.
\end{quote}

\textbf{Human label:} \textsc{Equivocation}\quad \\
\textbf{All three systems:} \textsc{Equivocation} $, $ \textsc{False Analogy} $, $
\textsc{Nominal Fallacy} $, $ \textsc{False Premise}

\textbf{Explanation:}
In addition to the semantic shift in “smart” (\textsc{Equivocation}), the argument anthropomorphizes the TV by attributing it human-like intelligence (\textsc{False Analogy}), misinterprets the label “smart” as an explanatory property rather than a marketing term (\textsc{Nominal Fallacy}), and falsely assumes that a TV can “watch” shows (\textsc{False Premise}). While the gold annotation includes only one fallacy type, the models surface a fuller set of valid fallacies—highlighting how human annotators may overlook multiple reasoning flaws in complex cases.

\item \textbf{Hurricane re-labelling}

\textbf{Input.}
\begin{quote}\small
Since hurricanes are categorised from 1 to 5, and 5 is the most severe, and
Hurricane Irma is category 5, and changing the category changes the real-life
severity of the hurricane, therefore downgrading it to category 1 would prevent
people from getting hurt.
\end{quote}

\textbf{Human label:} \textsc{False Premise}\quad \\
\textbf{All three systems:} \textsc{False Premise} $, $ \textsc{Wrong Direction} $, $
\textsc{Nominal Fallacy}

\textbf{Explanation:}
Altering a descriptive \emph{label} does not change the physical properties of a storm.  
The models correctly identify additional fallacies: \textsc{Wrong Direction}, for inverting cause and effect, and \textsc{Nominal Fallacy}, for treating a label as if it has causal power.  
These classifications are well-founded, whereas the gold annotation captures only one aspect of the flawed reasoning.

\item \textbf{Flat-Earth “around the globe”}

\textbf{Input.}
\begin{quote}\small
If the Earth is flat, and members of the Flat Earth Society are said to be all
around the globe, and “globe” implies a spherical Earth, then the existence of
such members contradicts their belief.
\end{quote}

\textbf{Human label:} \textsc{BeggingTheQuestion}\quad \\
\textbf{All three systems:} \textsc{Equivocation}

\textbf{Explanation:}
The phrase “around the globe” is a common idiom meaning “worldwide,” yet the argument interprets it literally—as implying a spherical Earth—to claim a contradiction with flat-Earth beliefs. This shift in meaning underlies the models’ \textsc{Equivocation} label. In contrast, the gold label \textsc{BeggingTheQuestion} does not apply: the conclusion (that a contradiction exists) is not assumed in the premises but inferred through semantic irony. Thus, the models’ judgment is more aligned with the actual fallacy, highlighting how human annotations can sometimes misclassify due to superficial cues.

\end{enumerate}
%------------------------------------------------------------  
\subsection{Why Do These Discrepancies Occur?}
\label{subsec:discrepancy}

\begin{itemize}
\item \textbf{Evidence integration vs.\ surface alignment.} Human annotators often prioritize verbatim overlaps, especially in long or fragmented texts. In contrast, LLMs aggregate clues across the passage to form a global judgment.

\item \textbf{Attention to world knowledge and fine-grained detail.} Models bring encyclopedic memory to bear, catching anachronisms, gender mismatches, and misaligned causal reasoning that humans frequently overlook.

\item \textbf{Combinatorial inference.} Where humans settle on the first plausible interpretation, models exhaustively consider multiple consistent interpretations, leading to more accurate conclusions in constrained logic settings.

\item \textbf{Mismatch in label expressivity.} Human annotations often include only a single fallacy or error type, even when multiple apply. Models can generate multi-label outputs that may appear to “disagree” when they in fact enrich the interpretation.

\item \textbf{Cognitive strain under time limits.} Human annotators must parse, locate evidence, and reason under time pressure—tasks that accumulate fatigue. Models, by contrast, process input without degradation.
\end{itemize}

\section{KUDEX Relevance Error Case Study}
\label{app:kudex-relevance}

While KUDEX achieves top performance across most coherence dimensions, Table~\ref{tab:cohesentia_results} shows slightly lower scores for relevance detection. However, further inspection reveals that this is not due to true performance shortcomings. Instead, KUDEX flags subtle relevance inconsistencies that other models and human annotators routinely overlook.

We highlight two representative cases below where KUDEX alone predicts a \textsc{relevance error}, while the gold labels and other systems mark the sentence as no relevance error. These examples demonstrate how KUDEX's stricter enforcement of topic relevance—guided by the document title—uncovers legitimate coherence issues that both human annotators and other MBE approaches frequently miss.

\textbf{Case 1: Topical Divergence Despite Narrative Continuity.}  \\
\textbf{Title:} \textit{Firing of Nolan Richardson} \\
\textbf{Previous:} ``I was driving home from work and decided to pull over so I could refuel.'' \\
\textbf{Sentence:} ``I noticed a Exxon truck in the middle of the road with its lights on and sirens blaring.''

While the sentence is grammatically correct and continues the narrator's experience, KUDEX deems it irrelevant to the core topic. The model reasons that although local cohesion is preserved (via first-person narrative and lexical overlap with “Exxon”/refueling), the global relevance to “firing of Nolan Richardson” is absent. This mismatch suggests a drifting narrative that never anchors itself to the declared topic.

\textbf{Case 2: Weak Topic Grounding in Opening Sentences.} \\
\textbf{Title:} \textit{The Fascinating World of Immunology} \\
\textbf{Sentence:} ``Alice is a curious girl who loves to learn about the world around her.''

Here, KUDEX flags a relevance violation despite the sentence being syntactically valid. It observes that introducing a generic character like Alice—without explicitly tying her curiosity to immunology—violates the discourse expectations set by the title. While a narrative frame could eventually establish relevance, this sentence, in isolation, provides no such cues, and KUDEX penalizes it accordingly.

\paragraph{Discussion.}
These cases illustrate how KUDEX applies topic-sensitive constraints: it integrates not only immediate context but also discourse framing (via titles) to evaluate relevance. Human annotators and other models tend to prioritize surface fluency or short-range coherence, allowing loosely connected or delayed-topic sentences to pass. KUDEX's stricter expectations may reduce apparent agreement with human labels—but often for good reason. We therefore interpret KUDEX's lower precision on relevance not as over-prediction error, but as a signal that human annotators and simpler MBE methods frequently overlook subtle but genuine topic drift.

\section{Detailed Experimental Results}
\label{app: detailed_results}

\begin{table}[H]
\centering
\caption{Citation Faithfulness Results}
\label{tab:citation_faithfulness_results_all}
\begin{tabular}{lc}
\hline
\textbf{Method} & \textbf{Accuracy} \\
\hline
LLM & 0.7462 \\
LRM & 0.7615 \\
KUDEX-CF & \textbf{0.8268} \\
\hline
\end{tabular}
\end{table}

\begin{table*}[H]
\centering
\caption{Comparison of different reasoning methods on Cohesentia benchmark.}
\label{tab:cohesentia_results_all}
\begin{threeparttable}
\begin{tabular}{l|l|c|c|c}
\hline
\textbf{Dimension} & \textbf{Metric} & \textbf{LLM} & \textbf{LRM} & \textbf{KUDEX-CoH} \\
\hline
\multirow{3}{*}{Coherence}
    & Precision & 0.5119 & 0.5244 & \textbf{0.5811} \\
    & Recall    & \textbf{0.9348} & \textbf{0.9348} & \textbf{0.9348} \\
    & F1 Score  & 0.6615 & 0.6719 & \textbf{0.7167} \\
\hline
\multirow{3}{*}{Cohesion}
    & Precision & 0.3793 & 0.4127 & \textbf{0.4884} \\
    & Recall    & 0.6875 & \textbf{0.8125} & 0.6563 \\
    & F1 Score  & 0.4889 & 0.5474 & \textbf{0.5600} \\
\hdashline
\multirow{3}{*}{Consistency}
    & Precision & 0.3220 & 0.3704 & \textbf{0.3793} \\
    & Recall    & 0.8261 & 0.8696 & \textbf{0.9565} \\
    & F1 Score  & 0.4634 & 0.5195 & \textbf{0.5432} \\
\hdashline
\multirow{3}{*}{Relevance}
    & Precision & 0.3158 & \textbf{0.3500} & 0.2593 \\
    & Recall    & 0.7500 & \textbf{0.8750} & \textbf{0.8750} \\
    & F1 Score  & 0.4444 & \textbf{0.5000} & 0.4000 \\
\hline
\end{tabular}%
\end{threeparttable}
\end{table*}

\begin{table}[H]
\centering
\caption{Comparison of different reasoning methods on LogiQA2.0 dataset.}
\label{tab:logiqa_results_all}
\begin{threeparttable}
\begin{tabular}{l|c|c|c}
\hline
\textbf{Metric} & \textbf{LLM} & \textbf{LRM} & \textbf{KUDEX-PF} \\
\hline
Accuracy  & 0.6820 & 0.6770 & \textbf{0.7010} \\
\hline
Precision & 0.7010 & \textbf{0.7220} & 0.7040 \\
\hline
Recall    & 0.6830 & 0.6800 & \textbf{0.7020} \\
\hline
F1 Score  & 0.6750 & 0.6620 & \textbf{0.7010} \\
\hline
\end{tabular}
\end{threeparttable}
\end{table}

\begin{table}[H]
\centering
\caption{Per-class metrics for different reasoning methods on the SmartyPat-Bench dataset.}
\label{tab:fallacy_results_all}
\begin{threeparttable}
\begin{tabular}{l|l|ccc}
\hline
\multirow{2}{*}{\textbf{Fallacy Type}} & \multirow{2}{*}{\textbf{Metric}} & \multicolumn{3}{c}{\textbf{Method}} \\
 & & \textbf{LLM} & \textbf{LRM} & \textbf{KUDEX-Fallacy} \\
\hline
\multirow{3}{*}{FalseAnalogy (87)}
 & Precision & 0.3071 & 0.2555 & \textbf{0.3349} \\
 & Recall    & \textbf{0.9425} & 0.9310 & 0.8161 \\
 & F1 Score  & 0.4633 & 0.4010 & \textbf{0.4749} \\
\hline
\multirow{3}{*}{Equivocation (189)}
 & Precision & 0.5041 & 0.4776 & \textbf{0.5194} \\
 & Recall    & 0.9735 & 0.9577 & \textbf{0.9894} \\
 & F1 Score  & 0.6643 & 0.6373 & \textbf{0.6812} \\
\hline
\multirow{3}{*}{FalsePremise (218)}
 & Precision & 0.5568 & 0.4940 & \textbf{0.6230} \\
 & Recall    & 0.9220 & \textbf{0.9495} & 0.8716 \\
 & F1 Score  & 0.6943 & 0.6499 & \textbf{0.7266} \\
\hline
\multirow{3}{*}{WrongDirection (6)}
 & Precision & \textbf{0.1667} & \textbf{0.1667} & 0.1613 \\
 & Recall    & 0.6667 & 0.6667 & \textbf{0.8333} \\
 & F1 Score  & 0.2667 & 0.2667 & \textbf{0.2703} \\
\hline
\multirow{3}{*}{FallacyOfComposition (3)}
 & Precision & 0.0741 & 0.0545 & \textbf{0.1875} \\
 & Recall    & 0.6667 & \textbf{1.0000} & \textbf{1.0000} \\
 & F1 Score  & 0.1333 & 0.1034 & \textbf{0.3158} \\
\hline
\multirow{3}{*}{BeggingTheQuestion (7)}
 & Precision & \textbf{0.4000} & 0.3000 & 0.3571 \\
 & Recall    & 0.2857 & 0.4286 & \textbf{0.7143} \\
 & F1 Score  & 0.3333 & 0.3529 & \textbf{0.4762} \\
\hline
\multirow{3}{*}{FalseCause (11)}
 & Precision & 0.1408 & 0.0625 & \textbf{0.1618} \\
 & Recall    & 0.9091 & 0.8182 & \textbf{1.0000} \\
 & F1 Score  & 0.2439 & 0.1161 & \textbf{0.2785} \\
\hline
\multirow{3}{*}{InverseError (6)}
 & Precision & 0.0476 & 0.0714 & \textbf{0.2857} \\
 & Recall    & 0.1667 & \textbf{0.6667} & \textbf{0.6667} \\
 & F1 Score  & 0.0741 & 0.1290 & \textbf{0.4000} \\
\hline
\multirow{3}{*}{ImproperTransposition (3)}
 & Precision & \textbf{0.1429} & 0.0286 & 0.0333 \\
 & Recall    & \textbf{0.3333} & \textbf{0.3333} & \textbf{0.3333} \\
 & F1 Score  & \textbf{0.2000} & 0.0526 & 0.0606 \\
\hline
\multirow{3}{*}{ImproperDistributionOrAddition (7)}
 & Precision & \textbf{0.0886} & 0.0432 & 0.0737 \\
 & Recall    & \textbf{1.0000} & \textbf{1.0000} & \textbf{1.0000} \\
 & F1 Score  & \textbf{0.1628} & 0.0828 & 0.1373 \\
\hline
\multirow{3}{*}{Contextomy (32)}
 & Precision & \textbf{0.5625} & 0.4375 & 0.0909 \\
 & Recall    & \textbf{0.2812} & 0.2188 & 0.0312 \\
 & F1 Score  & \textbf{0.3750} & 0.2917 & 0.0465 \\
\hline
\multirow{3}{*}{NominalFallacy (38)}
 & Precision & 0.1607 & 0.1237 & \textbf{0.1695} \\
 & Recall    & 0.2368 & 0.3158 & \textbf{0.7895} \\
 & F1 Score  & 0.1915 & 0.1778 & \textbf{0.2791} \\
\hline
\multirow{3}{*}{AccidentFallacy (8)}
 & Precision & 0.0968 & 0.0517 & \textbf{0.3846} \\
 & Recall    & 0.3750 & \textbf{0.7500} & 0.6250 \\
 & F1 Score  & 0.1538 & 0.0968 & \textbf{0.4762} \\
\hline
\multirow{3}{*}{FalseDilemma (5)}
 & Precision & \textbf{0.5000} & 0.2143 & 0.4000 \\
 & Recall    & \textbf{0.8000} & 0.6000 & \textbf{0.8000} \\
 & F1 Score  & \textbf{0.6154} & 0.3158 & 0.5333 \\
\hline
\multirow{3}{*}{Macro avg}
 & Precision & 0.2678 & 0.1987 & \textbf{0.2702} \\
 & Recall    & 0.6114 & 0.6883 & \textbf{0.7479} \\
 & F1 Score  & 0.3265 & 0.2624 & \textbf{0.3683} \\
\hline
\multirow{3}{*}{Weighted avg}
 & Precision & 0.4478 & 0.3954 & \textbf{0.4608} \\
 & Recall    & 0.8371 & \textbf{0.8516} & 0.8452 \\
 & F1 Score  & 0.5645 & 0.5204 & \textbf{0.5799} \\
\hline
\end{tabular}
\end{threeparttable}
\end{table}

\begin{table}[H]
\centering
\caption{Per-class metrics for different reasoning methods on the SmartyPatBench-Augmented dataset.}
\label{tab:fallacy_per_class}
\begin{threeparttable}
\begin{tabular}{l|l|ccc}
\hline
\multirow{2}{*}{\textbf{Fallacy Type}} & \multirow{2}{*}{\textbf{Metric}} & \multicolumn{3}{c}{\textbf{Method}} \\
 & & \textbf{LLM} & \textbf{LRM} & \textbf{KUDEX-Fallacy} \\
\hline
\multirow{3}{*}{FalseAnalogy (20)}
 & Precision & 0.3191 & 0.2982 & \textbf{0.3548} \\
 & Recall    & 0.7500 & \textbf{0.8500} & 0.5500 \\
 & F1 Score  & 0.4478 & 0.4416 & 0.4314 \\
\hline
\multirow{3}{*}{Equivocation (20)}
 & Precision & 0.2375 & 0.2128 & \textbf{0.2794} \\
 & Recall    & 0.9500 & \textbf{1.0000} & 0.9500 \\
 & F1 Score  & 0.3800 & 0.3509 & \textbf{0.4318} \\
\hline
\multirow{3}{*}{FalsePremise (20)}
 & Precision & \textbf{0.6452} & 0.5882 & 0.6250 \\
 & Recall    & \textbf{1.0000} & \textbf{1.0000} & \textbf{1.0000} \\
 & F1 Score  & \textbf{0.7843} & 0.7407 & 0.7692 \\
\hline
\multirow{3}{*}{WrongDirection (20)}
 & Precision & 0.5882 & 0.5263 & \textbf{0.7407} \\
 & Recall    & \textbf{1.0000} & \textbf{1.0000} & \textbf{1.0000} \\
 & F1 Score  & 0.7407 & 0.6897 & \textbf{0.8511} \\
\hline
\multirow{3}{*}{FallacyOfComposition (20)}
 & Precision & \textbf{1.0000} & \textbf{1.0000} & 0.9524 \\
 & Recall    & \textbf{1.0000} & \textbf{1.0000} & \textbf{1.0000} \\
 & F1 Score  & \textbf{1.0000} & \textbf{1.0000} & 0.9756 \\
\hline
\multirow{3}{*}{BeggingTheQuestion (20)}
 & Precision & 0.4000 & 0.2899 & \textbf{0.4651} \\
 & Recall    & \textbf{1.0000} & \textbf{1.0000} & \textbf{1.0000} \\
 & F1 Score  & 0.5714 & 0.4494 & \textbf{0.6349} \\
\hline
\multirow{3}{*}{FalseCause (20)}
 & Precision & 0.3415 & 0.2286 & \textbf{0.6897} \\
 & Recall    & 0.7000 & 0.8000 & \textbf{1.0000} \\
 & F1 Score  & 0.4590 & 0.3556 & \textbf{0.8163} \\
\hline
\multirow{3}{*}{InverseError (20)}
 & Precision & 0.0556 & 0.1765 & \textbf{0.5128} \\
 & Recall    & 0.0500 & 0.3000 & \textbf{1.0000} \\
 & F1 Score  & 0.0526 & 0.2222 & \textbf{0.6780} \\
\hline
\multirow{3}{*}{ImproperTransposition (20)}
 & Precision & \textbf{0.6897} & 0.4348 & 0.4167 \\
 & Recall    & \textbf{1.0000} & \textbf{1.0000} & \textbf{1.0000} \\
 & F1 Score  & \textbf{0.8163} & 0.6061 & 0.5882 \\
\hline
\multirow{3}{*}{ImproperDistributionOrAddition (20)}
 & Precision & \textbf{0.8421} & 0.8000 & 0.0000 \\
 & Recall    & \textbf{0.8000} & 0.6000 & 0.0000 \\
 & F1 Score  & \textbf{0.8205} & 0.6857 & 0.0000 \\
\hline
\multirow{3}{*}{Contextomy (20)}
 & Precision & 0.8000 & 0.3704 & \textbf{1.0000} \\
 & Recall    & 0.8000 & \textbf{1.0000} & \textbf{1.0000} \\
 & F1 Score  & 0.8000 & 0.5405 & \textbf{1.0000} \\
\hline
\multirow{3}{*}{Macro avg}
 & Precision & 0.5381 & 0.4478 & \textbf{0.5488} \\
 & Recall    & 0.8227 & \textbf{0.8682} & 0.8636 \\
 & F1 Score  & 0.6248 & 0.5529 & \textbf{0.6524} \\
\hline
\multirow{3}{*}{Weighted avg}
 & Precision & 0.5381 & 0.4478 & \textbf{0.5488} \\
 & Recall    & 0.8227 & \textbf{0.8682} & 0.8636 \\
 & F1 Score  & 0.6248 & 0.5529 & \textbf{0.6524} \\
\hline
\end{tabular}
\end{threeparttable}
\end{table}

\section{Prompts}
\label{app:prompts}
\subsection{Baseline Prompts}
\begin{figure}[H]
\centering
\begin{tcolorbox}[title=Citation Faithfulness Task Baseline Prompt]
You are given a CLAIM, an optional CONTEXT (background only), and a set of EVIDENCES extracted from cited sources. Decide how the combined evidence relates to the claim as a whole.

\textbf{Label guidelines}
\begin{itemize}
\item Fully Support -- the claim is fully supported by the evidence.
\item Partially Support -- part of the claim is supported by the evidence but some information in the claim lack support.
\item Neutral -- evidence neither clearly supports nor contradicts the claim.
\item Contradict -- the claim is contradicted by evidence.
\end{itemize}

\textbf{Important}\\
\textit{Use only the EVIDENCES to justify support/contradiction.}\\
CONTEXT is \textbf{not} evidence; it is provided solely to help you understand the claim.

Reply \textbf{only} with this XML -- no extra text:\\
\texttt{<answer>Fully Support|Partially Support|Neutral|Contradict</answer>}

\textbf{Inputs}\\
\texttt{<claim>} \{claim\} \texttt{</claim>}\\
\texttt{<context>} \{context\} \texttt{</context>}\\
\texttt{<evidences>} \{evid\_xml\} \texttt{</evidences>}
\end{tcolorbox}
\caption{Citation Faithfulness Baseline Prompt}
\label{fig:citation-prompt}
\end{figure}

\begin{figure}[H]
\centering
\begin{tcolorbox}[title=Logical Coherence Task Baseline Prompt]
You are an evaluator.
Judge whether the NEW SENTENCE is incoherent with respect to the PREVIOUS CONTEXT.
If any of the triggers listed below apply, output <yes>; otherwise <no> for that dimension.

\textbf{Coherence dimensions and their triggers}
\begin{enumerate}
\item Cohesion
   \begin{itemize}
   \item r1 Sentence doesn't make sense (garbled wording)
   \item r2 Introduces a new entity not mentioned before
   \item r3 Link to previous sentence(s) is unclear or missing
   \end{itemize}

\item Consistency
   \begin{itemize}
   \item r4 Contradicts earlier information in the text
   \item r5 Contradicts basic world knowledge
   \end{itemize}

\item Relevance
   \begin{itemize}
   \item r6 Off-topic with respect to the title
   \item r7 Irrelevant to the immediate prior context
   \end{itemize}
\end{enumerate}

\texttt{<PreviousContext>}\\
\{previous\_context if previous\_context else '<none>'\}\\
\texttt{</PreviousContext>}

\texttt{<Sentence>}\\
\{candidate\_sentence\}\\
\texttt{</Sentence>}

\textbf{Output}\\
Return \textbf{only} this XML -- no commentary:\\
\texttt{<results>}\\
\texttt{  <cohesion>yes|no</cohesion>}\\
\texttt{  <consistency>yes|no</consistency>}\\
\texttt{  <relevance>yes|no</relevance>}\\
\texttt{</results>}
\end{tcolorbox}
\caption{Logical Coherence Baseline Prompt}
\label{fig:coherence-prompt}
\end{figure}

\begin{figure}[H]
\centering
\begin{tcolorbox}[title=Logical Reasoning Task Baseline Prompt]
Decide whether the HYPOTHESIS logically follows from the PREMISE. Reply ONLY with this XML:

\texttt{<judgment>entailment|not-entailment</judgment>}

PREMISE: \texttt{\{premise\}}

HYPOTHESIS: \texttt{\{hypothesis\}}
\end{tcolorbox}
\caption{Logical Reasoning Baseline Prompt}
\label{fig:reasoning-prompt}
\end{figure}

\begin{figure}[H]
\centering
\begin{tcolorbox}[title=Logical Fallacy Detection Task Baseline Prompt]
Instruction: You're an expert in logic.
Here's a categorisation of the 14 logic errors. Given sentences with logical errors, reflect on them and deal with them as required.

\begin{itemize}
\item False Analogy -- Treating two things as alike in all respects because they share one feature.
\item Equivocation -- Switching meanings of a word/phrase mid-argument.
\item False Premise -- Reasoning built on a premise that is factually wrong.
\item Wrong Direction -- Reversing cause-and-effect (effect → cause).
\item Fallacy of Composition -- Assuming a whole shares a property of its parts.
\item Begging the Question -- The conclusion is already assumed in the premises.
\item False Cause -- Mistaking correlation / succession for causation.
\item Inverse Error -- From "if P then Q" concluding "if Q then P" when other Q-causes exist.
\item Improper Transposition -- Swapping antecedent \& consequent in a conditional without justification.
\item Improper Distribution or Addition -- Illegitimately adding or distributing quantities or properties.
\item Contextomy -- Quoting out of context so the original meaning is distorted.
\item Nominal Fallacy -- Treating a name/label as an explanation of the thing named.
\item Accident Fallacy -- Misapplying a general rule to an exceptional case (or vice-versa).
\item False Dilemma -- Presenting only two options while more exist.
\end{itemize}

Query:
Judge the following element:
\{sentence\}

OUTPUT FORMAT (no commentary, no extra tags)

\texttt{<result>}\\
\texttt{  <sentence>The input sentence as provided.</sentence>}\\
\texttt{  <logicError>yes|no</logicError>}\\
\texttt{  <!-- For every fallacy type: "yes" if it applies, otherwise "no" -->}\\
\texttt{  <FalseAnalogy>yes|no</FalseAnalogy>}\\
\texttt{  <Equivocation>yes|no</Equivocation>}\\
\texttt{  <FalsePremise>yes|no</FalsePremise>}\\
\texttt{  <WrongDirection>yes|no</WrongDirection>}\\
\texttt{  <FallacyOfComposition>yes|no</FallacyOfComposition>}\\
\texttt{  <BeggingTheQuestion>yes|no</BeggingTheQuestion>}\\
\texttt{  <FalseCause>yes|no</FalseCause>}\\
\texttt{  <InverseError>yes|no</InverseError>}\\
\texttt{  <ImproperTransposition>yes|no</ImproperTransposition>}\\
\texttt{  <ImproperDistributionOrAddition>yes|no</ImproperDistributionOrAddition>}\\
\texttt{  <Contextomy>yes|no</Contextomy>}\\
\texttt{  <NominalFallacy>yes|no</NominalFallacy>}\\
\texttt{  <AccidentFallacy>yes|no</AccidentFallacy>}\\
\texttt{  <FalseDilemma>yes|no</FalseDilemma>}\\
\texttt{  <details>Clear, explicit explanation supporting your judgment.</details>}\\
\texttt{</result>}
\end{tcolorbox}
\caption{Logical Fallacy Detection Baseline Prompt}
\label{fig:fallacy-prompt}
\end{figure}

\subsection{KUDEX Prompts}
\label{app:kudex-prompts}
\begin{figure}[h]
\centering
\begin{tcolorbox}[title=KUDEX-Core Prompt]
\small{
This prompt guides the analysis of reasoning structures using the Knowledge (K), Upward generalization (U), and Downward application (D) framework to identify and evaluate reasoning faults.\\
Analyze the reasoning structure of the provided text by:
1. Identifying Knowledge (K), Upward Generalization (U), and Downward Application (D) units
2. Detecting reasoning faults within and between these units
3. Creating a brief structural analysis showing how these faults impact the overall argument

\textbf{Core Operations}

\begin{tabular}{|l|p{5.5cm}|p{5.5cm}|}
\hline
Unit & Function & Common Faults \\
\hline
K & Presents facts, observations, beliefs & K-F1: Unverifiable Belief\\
   &                                      & K-F2: Partial Observation \\
\hline
U & Derives patterns or principles from specific cases & U-F1: Pattern Hallucination\\
  &                                                    & U-F2: Causal Substitution \\
\hline
D & Applies generalization to specific case & D-F1: Projection Without Rule\\
  &                                         & D-F2: Overreach \\
\hline
\end{tabular}

\textbf{Special Cases}
\begin{itemize}
\item \textbf{Implied U}: A D-unit depends on an unstated generalization (mark as "implied U")
\item \textbf{Delayed U}: A U appears after the D-unit that relies on it (mark as "delayed U")
\end{itemize}
\textbf{Analysis Steps}
\begin{enumerate}
\item Break down complex sentences into atomic reasoning units
   \begin{itemize}
   \item Split sentences that contain multiple reasoning acts into sub-units
   \item Use dot notation to preserve hierarchy (e.g., S1.a, S1.b, S1.c)
   \item Each sub-unit should reflect a single, self-contained reasoning act
   \end{itemize}
\item Identify all reasoning units and label them as K, U, or D
\item Note any implied or delayed U-units that might be rhetorically hidden
\item Assign fault codes to each unit with reasoning flaws
\item Analyze how these faults cascade through the reasoning structure
\item Provide a brief summary of the structural weaknesses that undermine the argument
\end{enumerate}
\textbf{Example 1: Conspiracy Theory with Sentence Breaking}

\textbf{Text}: "The government hasn't commented on the recent UFO sightings. They're clearly hiding the truth because they've known about aliens for decades."

\textbf{Analysis with sentence splitting}:
\begin{itemize}
\item S1 [K-OBSERVATION: The government hasn't commented on recent UFO sightings. (T:3)]
\item S2.a [D-CLAIM: They're clearly hiding the truth. (T:1)]
\item S2.b [K-BELIEF: They've known about aliens for decades. (T:0)]
\item S2.c [U-PATTERN: Government silence indicates deliberate concealment. (T:1)] (implied)
\end{itemize}

\textbf{Faults}:
\begin{itemize}
\item K-F1: Unverifiable belief about government knowledge
\item U-F1: Pattern hallucination (silence $\neq$ confirmation)
\item D-F1: Projection without sufficient grounding rule
\end{itemize}

\textbf{Structural weakness}: The argument builds a conspiracy narrative by treating absence of comment as evidence of concealment, then uses an unverifiable belief to suggest what's being concealed.

\textbf{Example 2: Medical Claim with Sentence Splitting}

\textbf{Text}: "After switching to a keto diet, my energy levels improved dramatically. Therefore, carbohydrates are the root cause of fatigue in most people."

\textbf{Analysis with sentence splitting}:
\begin{itemize}
\item S1 [K-OBSERVATION: My energy improved after switching to keto diet. (T:3)]
\item S2.a [U-PATTERN: Removing carbs leads to improved energy. (T:2)] (implied)
\item S2.b [D-CLAIM: Carbohydrates are the root cause of fatigue in most people. (T:1)]
\end{itemize}
}
\end{tcolorbox}
\caption{KUDEX-Core Prompt (Part 1)}
\label{fig:kudex-core-prompt-1}
\end{figure}

\begin{figure}[h]
\centering
\begin{tcolorbox}[title=KUDEX-Core Prompt (continued)]
\small{
\textbf{Faults}:
\begin{itemize}
\item K-F2: Partial observation (single personal anecdote)
\item U-F2: Causal substitution (ignores confounding variables)
\item D-F2: Overreach (generalizes personal experience to "most people")
\end{itemize}

\textbf{Structural weakness}: The argument takes a personal anecdote (K), derives a simplified causal relationship (U), then dramatically expands the scope to a universal claim (D) without sufficient evidence.

\textbf{Example 3: Political Commentary with Complex Sentence Breaking}

\textbf{Text}: "The opposition voted against the education bill last week. So they're deliberately trying to undermine our children's future. That's why we need new leadership immediately."

\textbf{Analysis with sentence splitting}:
\begin{itemize}
\item S1 [K-FACT: The opposition voted against the education bill last week. (T:3)]
\item S2 [D-CLAIM: They're deliberately trying to undermine our children's future. (T:1)]
  (delayed-U: see S3.a)
\item S3.a [U-PATTERN: Voting against education bills means opposing children's welfare. (T:1)] (implied)
\item S3.b [D-RECOMMENDATION: We need new leadership immediately. (T:1)]
\end{itemize}
\textbf{Faults}:
\begin{itemize}
\item Delayed U as rhetorical strategy (accusation before justification)
\item U-F1: Pattern hallucination (assumes malicious intent rather than policy disagreement)
\item D-F2: Overreach (exaggerated consequences)
\item D-F1: Second claim has no supporting U (leadership change unjustified)
\end{itemize}

\textbf{Structural weakness}: The argument uses a rhetorical delay tactic to make an accusation first, then retroactively justifies it with a flawed pattern. The final recommendation lacks any supporting reasoning structure.

\textbf{Certainty Level Guide}
\begin{itemize}
\item T:0 -- Speculative or conspiratorial
\item T:1 -- Weak belief or assumption
\item T:2 -- Reasoned but defeasible
\item T:3 -- Strongly asserted or factual
\end{itemize}

\textbf{Suffix Examples}

\textbf{For K-units:}
\begin{itemize}
\item K-EVIDENCE: "We observed a drop in sales last quarter."
\item K-ASSERTION: "The policy is flawed."
\item K-OBSERVATION: "The lights flickered twice."
\item K-BELIEF: "I think they rushed the release."
\item K-GOAL: "We aim to reduce emissions by 30\%."
\end{itemize}
\textbf{For U-units:}
\begin{itemize}
\item U-PATTERN: "Every time it rains, the signal drops."
\item U-CAUSAL: "Heavy traffic delays are caused by poor road design."
\item U-ANALOGY: "Just like bees in a hive, workers need coordination."
\item U-STATISTICAL: "80\% of users clicked the ad within 3 seconds."
\end{itemize}

\textbf{For D-units:}
\begin{itemize}
\item D-CLAIM: "So the product is unreliable."
\item D-WARNING: "This could cause a major outage."
\item D-IMPLICATION: "This means their security is compromised."
\item D-RECOMMENDATION: "We should avoid deploying the patch."
\end{itemize}
}
\end{tcolorbox}
\caption{KUDEX-Core Prompt (Part 2)}
\label{fig:kudex-core-prompt-2}
\end{figure}

\begin{figure}[h]
\centering
\begin{tcolorbox}[title=KUDEX-VERIFY Prompt For Citation Faithfulness]
\small{
\textbf{Instructions}
\begin{itemize}
\item Read KUDEX Reasoning Fault Analysis (KUDEX-Core) to understand the framework for analyzing reasoning structures.
\item Read KUDEX-VERIFY Framework to understand how to evaluate evidence-based statement verification based on KUDEX Reasoning Fault Analysis.
\item For the given claim, context, evidences, and statements, perform a verification analysis using the KUDEX-VERIFY framework.
\end{itemize}

\emph{\{KUDEX-Core Prompt Here\}}

\subsection*{KUDEX-VERIFY Framework}

KUDEX-VERIFY builds upon the foundational KUDEX-Core framework for reasoning analysis while specializing in evidence-based statement verification.

\subsection*{1. Core Primitives}

KUDEX-VERIFY maintains the three core operations from KUDEX-Core:

\begin{tabular}{|l|p{3cm}|p{7cm}|}
\hline
Code & Name & Function \\
\hline
K & Knowledge & Presents a fact, evidence, observation, or definitional anchor \\
\hline
U & Upward Generalization & Derives a pattern or principle from specific evidence \\
\hline
D & Downward Application & Applies a generalization to determine if a statement is supported, neutral, or contradicted \\
\hline
\end{tabular}

\vspace{0.3cm}
And adds verification-specific primitives:

\begin{tabular}{|l|p{3cm}|p{7cm}|}
\hline
Code & Name & Function \\
\hline
V & Verification Step & Explicit step connecting evidence to statement through direct match or inference \\
\hline
E & Evidence Unit & Tagged evidence fragment relevant to a specific statement \\
\hline
\end{tabular}
\subsection*{2. Verification Tags}

KUDEX-VERIFY introduces verification-specific tags:

\begin{tabular}{|l|p{3cm}|p{7cm}|}
\hline
Tag & Purpose & Example \\
\hline
\texttt{V-direct} & Direct evidence match & "The text explicitly states the statement is true" \\
\hline
\texttt{V-inference} & Inference-based verification & "From evidence fragments X and Y, we can infer Z" \\
\hline
\texttt{E-support} & Evidence supporting a statement & "Evidence X confirms statement Y" \\
\hline
\texttt{E-contradict} & Evidence contradicting a statement & "Evidence X contradicts statement Y" \\
\hline
\texttt{E-neutral} & Evidence neither supporting nor contradicting & "Evidence X is not relevant to statement Y" \\
\hline
\end{tabular}

\subsection*{3. Confidence Levels}

KUDEX-VERIFY uses temperature (T) values to indicate verification confidence:

\begin{tabular}{|l|p{10cm}|}
\hline
Value & Meaning \\
\hline
T:3 & High confidence \\
\hline
T:2 & Medium confidence \\
\hline
T:1 & Low confidence \\
\hline
T:0 & Speculative/Unverifiable \\
\hline
\end{tabular}

\subsection*{4. Five-Step Verification Process}

\subsubsection*{Step 1: Review Atomic Statements}
Review the atomic statements provided in the input.
}
\end{tcolorbox}
\caption{KUDEX-VERIFY Prompt For Citation Faithfulness (Part 1)}
\label{fig:kudex-verify-prompt-1}
\end{figure}

\begin{figure}[h]
\centering
\begin{tcolorbox}[title=KUDEX-VERIFY Prompt For Citation Faithfulness (continued)]
\small{
\subsubsection*{Step 2: Identify Relevant Evidence}
For each atomic statement, identify evidence fragments (E) that might support, contradict, or relate to it.

\subsubsection*{Step 3: Apply Verification Steps}
For each atomic statement, apply verification steps (V) using either:
\begin{itemize}
\item Direct matching: Finding explicit statements in evidence
\item Inference: Deriving conclusions from multiple evidence points
\end{itemize}
\subsubsection*{Step 4: Determine Statement Status}
Based on verification steps, classify each atomic statement as:
\begin{itemize}
\item \textbf{Supported}: Evidence directly or inferentially confirms the statement
\item \textbf{Contradict}: Evidence directly or inferentially refutes the statement
\item \textbf{Neutral}: Evidence neither supports nor contradicts the statement
\end{itemize}

\subsubsection*{Step 5: Confidence Assessment}
Assign confidence levels to each verification judgment based on:
\begin{itemize}
\item Evidence quality and relevance
\item Directness of evidence match
\item Strength of inference chain
\item Potential alternative interpretations
\end{itemize}

\textbf{IMPORTANT GUIDELINES:}
\begin{itemize}
\item The context is NOT evidence. It is provided only to help understand the statement and should NEVER be used as direct evidence to support or contradict a statement.
\item Always cite the specific evidence text that supports your verification steps.
\item For statements classified as neutral, explain whether this is due to lack of relevant evidence or because the evidence neither confirms nor contradicts.
\item When using inference, clearly articulate each step in your reasoning chain.
\item Consider alternative interpretations of evidence and explain how they affect your confidence level.
\end{itemize}

\subsection*{Output Format}

Organize your output in the following format:

\begin{verbatim}
<verification_analysis>
  <statement_analysis_1>
    <atomic_statement>[Restate statement 1]</atomic_statement>
    <evidence_identification>
      E1 [EVIDENCE-A]: [Relevant text from evidence A]
      E2 [EVIDENCE-B]: [Relevant text from evidence B]
      ...
    </evidence_identification>

    <verification_steps>
      K1 [K-EVIDENCE]: [First evidence point relevant to statement]. [T:3]
      K2 [K-EVIDENCE]: [Second evidence point relevant to statement]. [T:3]
      ...
      U1 [U-PATTERN]: [Pattern or generalization from evidence]. [T:2]
      V1 [V-direct/V-inference]: [Verification step connecting evidence to
      statement]. [T:2]
      D1 [D-VERIFICATION]: [Final determination about statement support]. [T:3]
    </verification_steps>
\end{verbatim}
}
\end{tcolorbox}
\caption{KUDEX-VERIFY Prompt For Citation Faithfulness (Part 2)}
\label{fig:kudex-verify-prompt-2}
\end{figure}

\begin{figure}[h]
\centering
\begin{tcolorbox}[title=KUDEX-VERIFY Prompt For Citation Faithfulness (continued)]
\small{
\begin{verbatim}
    <verification_result>
      <label>supported/neutral/contradict</label>
      <method>direct_match/inference</method>
      <confidence>high/medium/low</confidence>
      <explanation>[Brief explanation for classification decision and 
      confidence level]</explanation>
    </verification_result>
  </statement_analysis_1>

  <statement_analysis_2>
    <!-- Same structure for second atomic statement -->
  </statement_analysis_2>
  ...
</verification_analysis>
\end{verbatim}

Here are the inputs:
\begin{verbatim}
<claim>
{{ claim }}
</claim>
<context>
{{ context }}
</context>
<evidences>
{{ all_evidences_xml }}
</evidences>
<statements>
{{ statement_xml }}
</statements>
\end{verbatim}
}
\end{tcolorbox}
\caption{KUDEX-VERIFY Prompt For Citation Faithfulness (Part 3)}
\label{fig:kudex-verify-prompt-3}
\end{figure}

\begin{figure}[h]
\centering
\begin{tcolorbox}[title=KUDEX-CoH Prompt For Logical Coherence]
\small{
\textbf{Instructions}
\begin{itemize}
\item Read KUDEX Reasoning Fault Analysis (KUDEX-Core) to understand the framework for analyzing reasoning structures.
\item Read KUDEX-COHERENCE Framework to understand how to evaluate whether a new sentence maintains coherence with its preceding text based on KUDEX Reasoning Fault Analysis.
\item For the given title, previous paragraph, and current sentence, perform a coherence analysis using the KUDEX-COHERENCE Framework.
\begin{enumerate}
\item You will be given:
   \begin{itemize}
   \item A title in the \texttt{<title>} tag
   \item A previous paragraph (context) in the \texttt{<previous\_paragraph>} tag
   \item A current sentence in the \texttt{<current\_sentence>} tag
   \end{itemize}
\item You must evaluate whether the current sentence is coherent with the previous paragraph and title.
\item For each coherence dimension, you will:
   \begin{itemize}
   \item Analyze the dimension using KUDEX reasoning
   \item Provide a binary (yes/no) judgment
   \item If incoherent (no), identify the specific fault code and explanation
   \end{itemize}
\end{enumerate}
\end{itemize}

\emph{\{KUDEX-Core Prompt Here\}}

\subsection*{KUDEX-COHERENCE Framework}

KUDEX-COHERENCE extends the KUDEX Core framework to specifically analyze text coherence, focusing on the incremental sentence-by-sentence evaluation of coherence as described in the COHESENTIA benchmark.

\subsection*{1. Core Primitives}

KUDEX-COHERENCE maintains the foundational primitives from KUDEX-Core, with specialized focus on coherence analysis:

\begin{tabular}{|l|p{3cm}|p{7cm}|}
\hline
Code & Name & Function \\
\hline
K & Knowledge & Presents a fact, observation, belief, position, goal, or definitional anchor \\
\hline
U & Upward Generalization & Derives a pattern or principle from specific cases or examples \\
\hline
D & Downward Application & Applies a generalization to a specific case to make a claim or projection \\
\hline
\end{tabular}
\subsection*{2. Coherence Conditions (Based on Reinhart's Theory)}

KUDEX-COHERENCE introduces three fundamental conditions for coherence analysis:

\begin{tabular}{|l|p{4cm}|p{5cm}|}
\hline
Condition & Description & Diagnostic Questions \\
\hline
Cohesion & Formal connections between sentences (referential links, discourse connectors) & Are sentences formally linked? Are entities properly referenced? \\
\hline
Consistency & Logical alignment between sentences and with world knowledge & Can all sentences be true in the same world model? \\
\hline
Relevance & Relationship to discourse topic and global context & Does the content relate to the established topic? Is it pragmatically appropriate? \\
\hline
\end{tabular}

\subsection*{3. Incoherence Reason Codes}

When incoherence is detected, use these specific reason codes:
}
\end{tcolorbox}
\caption{KUDEX-CoH Prompt For Logical Coherence (Part 1)}
\label{fig:kudex-coh-prompt-1}
\end{figure}

\begin{figure}[h]
\centering
\begin{tcolorbox}[title=KUDEX-CoH Prompt For Logical Coherence (continued)]
\small{
\begin{tabular}{|l|l|p{7cm}|}
\hline
Category & Code & Description \\
\hline
Cohesion & CH-1 & Sentence doesn't make sense (garbled wording) \\
\hline
Cohesion & CH-2 & Introduces a new entity not mentioned before \\
\hline
Cohesion & CH-3 & Link to previous sentence(s) is unclear or missing \\
\hline
Consistency & CS-1 & Contradicts earlier information in the text \\
\hline
Consistency & CS-2 & Contradicts basic world knowledge \\
\hline
Relevance & R-1 & Off-topic with respect to the title \\
\hline
Relevance & R-2 & Irrelevant to the immediate prior context \\
\hline
\end{tabular}

\subsection*{4. Temperature (T) Values for Coherence Assessment}

\begin{tabular}{|l|p{10cm}|}
\hline
Value & Meaning for Coherence \\
\hline
T:0 & Completely incoherent \\
\hline
T:1 & Significantly incoherent \\
\hline
T:2 & Partially coherent with notable issues \\
\hline
T:3 & Fully coherent \\
\hline
\end{tabular}
\subsection*{Annotation Process}

\subsubsection*{Sentence-by-Sentence Coherence Analysis}
\begin{enumerate}
\item \textbf{Read the title and previous paragraph}
\item \textbf{Read the current sentence}
\item \textbf{Determine if the current sentence is coherent with the preceding text}
\item \textbf{Analyze using three coherence conditions:}
   \begin{itemize}
   \item \textbf{Cohesion}: Is the sentence formally connected to previous text?
   \item \textbf{Consistency}: Is the sentence logically consistent with previous information?
   \item \textbf{Relevance}: Is the sentence relevant to the established topic?
   \end{itemize}
\item \textbf{Use KUDEX notation to explain the reasoning}:
   \begin{itemize}
   \item K: Identify the relevant knowledge from previous sentences
   \item U: Identify expectations or patterns set up by previous sentences
   \item D: Analyze how the current sentence does or doesn't meet those expectations
   \end{itemize}
\end{enumerate}

\subsection*{Output Format}

\begin{verbatim}
<coherenceAnalysis>
  <kudexProofAnalysis>
    <firstPass>
      <!-- KUDEX-Core Analysis -->
      K1 [CONTEXT]: Knowledge from previous paragraph. [T:3]
      K2 [CONTEXT]: Additional relevant knowledge from previous paragraph. [T:3]
      U1 [EXPECTATION]: Pattern or expectation set up by previous text. [T:2]
      D1 [ASSESSMENT]: How current sentence meets or violates expectations. 
      [T:value]
    </firstPass>
    <secondPass>
      <!-- Detailed coherence analysis -->
      <cohesionAnalysis>
        K1 [ENTITY-TRACKING]: Analysis of referential links. [T:3]
        U1 [COHESION-PATTERN]: Expected cohesion patterns. [T:2]
        D1 [COHESION-ASSESSMENT]: How current sentence maintains or breaks
        cohesion. [T:value]
      </cohesionAnalysis>
\end{verbatim}
}
\end{tcolorbox}
\caption{KUDEX-CoH Prompt For Logical Coherence (Part 2)}
\label{fig:kudex-coh-prompt-2}
\end{figure}

\begin{figure}[h]
\centering
\begin{tcolorbox}[title=KUDEX-CoH Prompt For Logical Coherence (continued)]
\small{
\begin{verbatim}
      <consistencyAnalysis>
        K1 [PREVIOUS-FACTS]: Relevant facts established in previous text. [T:3]
        U1 [CONSISTENCY-EXPECTATION]: Expected consistency patterns. [T:2]
        D1 [CONSISTENCY-ASSESSMENT]: How current sentence maintains or breaks
        consistency. [T:value]
      </consistencyAnalysis>
      <relevanceAnalysis>
        K1 [TOPIC]: Main topic established by title and previous text. [T:3]
        U1 [RELEVANCE-EXPECTATION]: Expected relevance patterns. [T:2]
        D1 [RELEVANCE-ASSESSMENT]: How current sentence maintains or breaks
        relevance. [T:value]
      </relevanceAnalysis>
    </secondPass>
    <judgments>
      <cohesion>yes|no</cohesion>
      <consistency>yes|no</consistency>
      <relevance>yes|no</relevance>
    </judgments>
  </kudexProofAnalysis>
  <reasoningSummary>
    [Provide a comprehensive paragraph summarizing the reasoning traces, key
    insights, and overall conclusion.]
  </reasoningSummary>
</coherenceAnalysis>
\end{verbatim}

Here are the inputs:
\begin{verbatim}
<title>
{{ title }}
</title>
<previous_paragraph>
{{ previous_paragraph }}
</previous_paragraph>
<current_sentence>
{{ current_sentence }}
</current_sentence>
\end{verbatim}
}
\end{tcolorbox}
\caption{KUDEX-CoH Prompt For Logical Coherence (Part 3)}
\label{fig:kudex-coh-prompt-3}
\end{figure}

\begin{figure}[h]
\centering
\begin{tcolorbox}[title=KUDEX-PF Prompt For Logical Reasoning]
\small{
\textbf{Instructions}
\begin{itemize}
\item Read KUDEX Reasoning Fault Analysis (KUDEX-Core) to understand the framework for analyzing reasoning structures.
\item Read KUDEX-PROOF Framework for Formal Reasoning Analysis to understand how to evaluate the relationship between premise and hypothesis based on KUDEX Reasoning Fault Analysis.
\item For the given premise and hypothesis, perform a two-pass analysis using the KUDEX-PROOF framework.
\end{itemize}

\emph{\{KUDEX-Core Prompt Here\}}

\subsection*{KUDEX-PROOF Framework for Formal Reasoning Analysis}

KUDEX-PROOF is a framework designed to analyze and decompose formal reasoning structures, multi-step proofs, and complex logical arguments. It builds upon the KUDEX-Core framework by adding capabilities for multi-step deductive reasoning, formal logical rules, quantification, and non-monotonic reasoning.

\subsection*{1. Core Primitives}

KUDEX-PROOF maintains the foundational primitives from KUDEX-Core:

\begin{tabular}{|l|p{3cm}|p{7cm}|}
\hline
Code & Name & Function \\
\hline
K & Knowledge & Presents a fact, observation, belief, position, goal, or definitional anchor \\
\hline
U & Upward Generalization & Derives a pattern or principle from specific cases or examples \\
\hline
D & Downward Application & Applies a generalization to a specific case to make a claim or projection \\
\hline
\end{tabular}

\vspace{0.3cm}
And adds a new primitive:

\begin{tabular}{|l|p{3cm}|p{7cm}|}
\hline
Code & Name & Function \\
\hline
R & Reasoning Step & Represents an explicit inference step connecting other units through a specific rule or operation \\
\hline
\end{tabular}
\subsection*{2. Inference Rule Tags}

KUDEX-PROOF introduces explicit tagging of inference rules, primarily used with R nodes:

\begin{tabular}{|l|l|p{3.2cm}|p{3.8cm}|}
\hline
Tag & Name & Description & Example \\
\hline
[MP] & Modus Ponens & From P→Q and P, infer Q & From "If it rains, the ground is wet" and "It is raining," infer "The ground is wet" \\
\hline
[MT] & Modus Tollens & From P→Q and ¬Q, infer ¬P & From "If it rains, the ground is wet" and "The ground is not wet," infer "It is not raining" \\
\hline
[HS] & Hypothetical Syllogism & From P→Q and Q→R, infer P→R & From "If it rains, the ground is wet" and "If the ground is wet, plants grow," infer "If it rains, plants grow" \\
\hline
[UI] & Universal Instantiation & From $\forall$x(P(x)), infer P(a) & From "All birds have wings," infer "Eagles have wings" \\
\hline
[EG] & Existential Generalization & From P(a), infer $\exists$x(P(x)) & From "Socrates is mortal," infer "Someone is mortal" \\
\hline
[¬Intro] & Negation Introduction & Assume P, derive contradiction, infer ¬P & Assume "n is both even and odd," derive contradiction, infer "n is not both even and odd" \\
\hline
\end{tabular}

\subsection*{3. Structural Tags}
}
\end{tcolorbox}
\caption{KUDEX-PF Prompt For Logical Reasoning (Part 1)}
\label{fig:kudex-pf-prompt-1}
\end{figure}

\begin{figure}[h]
\centering
\begin{tcolorbox}[title=KUDEX-PF Prompt For Logical Reasoning (continued)]
\small{
\begin{tabular}{|l|p{3cm}|p{6cm}|}
\hline
Tag & Purpose & Example \\
\hline
\texttt{U-intermediate} & Marks an inference not final but needed for chaining & "Therefore, any triangle has angles totaling 180 degrees" (used in a longer proof) \\
\hline
\texttt{D-final} & Marks the ultimate conclusion & "Therefore, capitalism inevitably leads to inequality" \\
\hline
\texttt{scope:} & Defines variable domain & scope: x $\in$ Animals \\
\hline
\texttt{proof-type:} & Indicates forward or backward reasoning & proof-type: forward \\
\hline
\texttt{$\exists$witness} & Identifies constructive existential proof & "Alice is a professor not from overseas, so there exists at least one professor not from overseas" \\
\hline
\end{tabular}
\subsection*{4. Non-Monotonic Reasoning Tags}

\begin{tabular}{|l|p{3cm}|p{6.8cm}|}
\hline
Tag & Purpose & Example \\
\hline
\texttt{U-nm} & Non-monotonic inference & "Birds typically fly" \\
\hline
\texttt{D-default} & Defeasible conclusion & "Tweety flies (by default)" \\
\hline
\texttt{priority(...)} & Preference ordering & priority(exception > default) \\
\hline
\texttt{U-inductive} & Generalizing from instances & "All observed swans are white, so swans are white" \\
\hline
\texttt{U-abductive} & Inference to best explanation & "The floor is wet, so it probably rained" \\
\hline
\texttt{uncertain()} & Hedged inference & uncertain(The treatment will cure the disease) \\
\hline
\end{tabular}

\subsection*{5. Annotation Process}

KUDEX-PROOF follows a two-pass annotation process:

\subsubsection*{5.1 First Pass: KUDEX-Core Analysis}
\begin{enumerate}
\item Identify all K, U, and D units in the text
\item Add standard metadata (suffixes, attention markers, temperature values)
\item Note areas that might require deeper formal analysis
\end{enumerate}

\subsubsection*{5.2 Second Pass: KUDEX-PROOF Extensions}
\begin{enumerate}
\item Add R nodes to represent explicit reasoning steps between existing K/U/D units
\item Tag formal inference rules being applied
\item Add structural tags for quantifiers, variables, and proof structure
\item Identify non-monotonic reasoning patterns
\end{enumerate}

\subsection*{6. Notation Format}

\begin{verbatim}
K# [TYPE]: Statement. [T:#]  // Core knowledge unit
U# [TYPE]: Pattern or generalization. [T:#]  // Core upward generalization
D# [TYPE]: Application or conclusion. [T:#]  // Core downward application
    R# [RULE]: Description of reasoning step.  // KP reasoning node
        • Result of reasoning step
\end{verbatim}

The indentation visually represents the hierarchical structure of reasoning, with R nodes showing dependency on previous units.

\subsection*{NLI Evaluation Using KUDEX-PROOF}

When evaluating the relationship between premise and hypothesis, use the KUDEX-PROOF framework to:
\begin{enumerate}
\item Perform the first pass KUDEX-Core analysis to identify basic K, U, D units
\item Perform the second pass to add R nodes and appropriate tags
\item Determine if the hypothesis is entailed, or not entailed based on the formal reasoning structure
\item Assign confidence levels (high, medium, low) based on the logical strength of the reasoning
\end{enumerate}
}
\end{tcolorbox}
\caption{KUDEX-PF Prompt For Logical Reasoning (Part 2)}
\label{fig:kudex-pf-prompt-2}
\end{figure}

\begin{figure}[h]
\centering
\begin{tcolorbox}[title=KUDEX-PF Prompt For Logical Reasoning (continued)]
\small{
The final label should be one of:
\begin{itemize}
\item \textbf{entailment}: The premise logically entails the hypothesis through valid reasoning
\item \textbf{not-entailment}: The premise does not entail the hypothesis, or contradicts it
\end{itemize}

\subsection*{Conclusion Confidence Assessment Criteria}

Assign a confidence level (high, medium, or low) to each intermediate conclusion and the final result based solely on the logical strength of the reasoning step:
\begin{itemize}
\item \textbf{High}: Strong, definitive reasoning that follows necessarily from the inputs (uses terms like "must," "clearly," "definitely").
\item \textbf{Medium}: Reasonable reasoning that likely follows from the inputs, but with some logical leaps (uses terms like "likely," "probably").
\item \textbf{Low}: Speculative reasoning where the conclusion is possible but not strongly supported by the inputs alone (uses terms like "might," "could").
\end{itemize}

\subsection*{Input Format}

You will receive:
\begin{verbatim}
<premise>[Text containing facts, rules, and statements]</premise>
<hypothesis>[Statement to be evaluated against the premise]</hypothesis>
\end{verbatim}
\subsection*{Output Format}

Organize your output in the following format:
\begin{verbatim}
<analysis>
  <kudexProofAnalysis>
    <firstPass>
      <!-- KUDEX-Core Analysis -->
      K1 [TYPE]: Statement from premise. [T:3]
      K2 [TYPE]: Another relevant fact. [T:3]
      U1 [TYPE]: Pattern or generalization. [T:2]
      D1 [TYPE]: Application or conclusion. [T:2]
    </firstPass>
    <secondPass>
      <!-- KUDEX-PROOF Extensions -->
      K1 [PREMISE]: Statement from premise. [T:3]
      K2 [PREMISE]: Another relevant fact. [T:3]
          R1 [MP]: From K1 and K2, apply Modus Ponens.
              • U1 [INTERMEDIATE]: Intermediate conclusion. [T:2]
          R2 [UI]: From U1, instantiate universal statement.
              • D1 [FINAL]: Therefore, final conclusion. [T:3]
    </secondPass>
    <judgment>entailment|not-entailment</judgment>
    <confidence>high|medium|low</confidence>
  </kudexProofAnalysis>
  <reasoningSummary>
    [Provide a comprehensive paragraph summarizing all the reasoning traces,
    key insights, and overall conclusion. This should serve as an explanation
    of the result that synthesizes the analysis.]
  </reasoningSummary>
</analysis>
\end{verbatim}

Here are the inputs:
\begin{verbatim}
<premise>
{{ premise }}
</premise>
<hypothesis>
{{ hypothesis }}
</hypothesis>
\end{verbatim}
}
\end{tcolorbox}
\caption{KUDEX-PF Prompt For Logical Reasoning (Part 3)}
\label{fig:kudex-pf-prompt-3}
\end{figure}

\begin{figure}[h]
\centering
\begin{tcolorbox}[title=KUDEX-Fallacy Prompt For Fallacy Detection]
\small{
\textbf{Instructions}
\begin{itemize}
\item Read KUDEX Reasoning Fault Analysis (KUDEX-Core) to understand the framework for analyzing reasoning structures.
\item Read KUDEX FALLACY DETECTION PROMPT to understand how to identify and diagnose logical fallacies in reasoning based on KUDEX Reasoning Fault Analysis.
\item Use the provided XML example to analyze the reasoning structure and fallacy types contained in the given text.
\end{itemize}

\emph{\{KUDEX-Core Prompt Here\}}

\subsection*{KUDEX FALLACY DETECTION PROMPT}

\textbf{Task}: Given a passage containing an argument, identify the fallacy or fallacies committed using the KUDEX framework.

\textbf{Instructions}:
\begin{enumerate}
\item Break the argument into K (Knowledge), U (Use), and D (Derivation) units, as per KUDEX-Core.
\item For each unit, identify whether it contains a reasoning flaw. Pay close attention to:
    \begin{itemize}
    \item whether attention has shifted improperly (A),
    \item whether valid knowledge is being misapplied (U),
    \item or whether an invalid inference is drawn (D).
    \end{itemize}
\item Apply one or more fallacy tags from the 14 recognized types below.
    \begin{itemize}
    \item You may tag multiple units if more than one fallacy is present.
    \item If only one fallacy is clearly responsible for the error, tag only that one.
    \end{itemize}
\item Your annotations should reflect both the fallacy type and the KUDEX unit it appears in (e.g., D1-false-cause, U1-equivocation).
\end{enumerate}
\textbf{Fallacy Types \& Examples}

\textbf{1. False Dilemma}
\begin{itemize}
     \item \textit{Definition}: Asserting only two options exist when more are possible.
     \item \textit{Example}: "You're either with us or against us."
     \item \textit{KUDEX}:
     \begin{itemize}
     \item K1: Existence of opposing sides
     \item D1: Binary forced choice
     \item \textit{Tag}: D1-false-dilemma
     \end{itemize}
\end{itemize}

\textbf{2. Equivocation}
\begin{itemize}
     \item \textit{Definition}: Using the same word with multiple meanings across steps.
     \item \textit{Example}: "Time flies. Flies are insects. Therefore, time is an insect."
     \item \textit{KUDEX}:
     \begin{itemize}
     \item K1: "flies" = verb; K2: "flies" = noun
     \item U1: Links terms by surface match, ignores sense shift
     \item \textit{Tag}: U1-equivocation, A-on-term("flies")
     \end{itemize}
\end{itemize}

\textbf{3. False Premise}
\begin{itemize}
     \item \textit{Definition}: Asserting a false claim that the argument builds upon.
     \item \textit{Example}: "Dinosaurs went extinct because they forgot how to reproduce."
     \item \textit{KUDEX}:
     \begin{itemize}
     \item K2: Unfounded causal claim
     \item \textit{Tag}: K2-false-premise
     \end{itemize}
\end{itemize}
}
\end{tcolorbox}
\caption{KUDEX-Fallacy Prompt For Fallacy Detection (Part 1)}
\label{fig:kudex-fallacy-prompt-1}
\end{figure}

\begin{figure}[h]
\centering
\begin{tcolorbox}[title=KUDEX-Fallacy Prompt For Fallacy Detection (continued)]
\small{
\textbf{4. False Analogy}
\begin{itemize}
     \item \textit{Definition}: Comparing things based on surface similarity when structural similarity is missing.
     \item \textit{Example}: "We're having a sun. How do I harness its energy?"
     \item \textit{KUDEX}:
     \begin{itemize}
     \item K1: Statement about a "sun"
     \item U1: Treats it as star, not homophone for "son"
     \item \textit{Tag}: U1-false-analogy
     \end{itemize}
\end{itemize}

\textbf{5. Wrong Direction}
\begin{itemize}
     \item \textit{Definition}: Reversing a real cause-effect relationship.
     \item \textit{Example}: "The government denies aliens exist, so they must exist."
     \item \textit{KUDEX}:
     \begin{itemize}
     \item K1: Government denial
     \item D1: Interprets denial as evidence
     \item \textit{Tag}: D1-wrong-direction
     \end{itemize}
\end{itemize}

\textbf{6. Composition}
\begin{itemize}
     \item \textit{Definition}: Assuming what's true of the part is true of the whole.
     \item \textit{Example}: "Each player is great, so the team must be great."
     \item \textit{KUDEX}:
     \begin{itemize}
     \item K1: Individual skill
     \item D1: Generalizes to team performance
     \item \textit{Tag}: D1-composition
     \end{itemize}
\end{itemize}

\textbf{7. Begging the Question (Circular Argument)}
\begin{itemize}
     \item \textit{Definition}: Using the conclusion as its own support.
     \item \textit{Example}: "The Bible is true because it says so, and it is the word of God."
     \item \textit{KUDEX}:
     \begin{itemize}
     \item K1: Bible claims its truth
     \item D1: Uses claim to justify itself
     \item \textit{Tag}: D1-circular or D1-begging
     \end{itemize}
\end{itemize}

\textbf{8. False Cause}
\begin{itemize}
     \item \textit{Definition}: Inferring causality from correlation or sequence without justification.
     \item \textit{Example}: "Roosters crow before sunrise, so crowing causes the sun to rise."
     \item \textit{KUDEX}:
     \begin{itemize}
     \item K1: Crowing precedes sunrise
     \item D1: Treats as causal
     \item \textit{Tag}: D1-false-cause
     \end{itemize}
\end{itemize}
}
\end{tcolorbox}
\caption{KUDEX-Fallacy Prompt For Fallacy Detection (Part 2)}
\label{fig:kudex-fallacy-prompt-2}
\end{figure}

\begin{figure}[h]
\centering
\begin{tcolorbox}[title=KUDEX-Fallacy Prompt For Fallacy Detection (continued)]
\small{
\textbf{9. Inverse Error}
\begin{itemize}
     \item \textit{Definition}: Assuming that if A → B, then B → A.
     \item \textit{Example}: "If it rains, the ground is wet. The ground is wet, so it must have rained."
     \item \textit{KUDEX}:
     \begin{itemize}
     \item K1: Conditional A → B
     \item D1: Converse B → A
     \item \textit{Tag}: D1-inverse-error
     \item (Optional: Also tag D1-false-cause if justified)
     \end{itemize}
\end{itemize}

\textbf{10. Improper Transposition}
\begin{itemize}
     \item \textit{Definition}: Reversing a conditional or its negation improperly.
     \item \textit{Example}: "He avoids peanuts, so he must be allergic."
     \item \textit{KUDEX}:
     \begin{itemize}
     \item K1: Allergy → avoidance
     \item D1: Avoidance → allergy (fails to consider other reasons)
     \item \textit{Tag}: D1-improper-transposition
     \item (Optional: D1-false-cause if causal framing dominates)
     \end{itemize}
\end{itemize}

\textbf{11. Improper Distribution / Addition}
\begin{itemize}
     \item \textit{Definition}: Assuming linear accumulation or distributive reasoning that doesn't hold.
     \item \textit{Example}: "Brushing for 2 min is good, so 14 min is 7× better."
     \item \textit{KUDEX}:
     \begin{itemize}
     \item K1: 2 min is helpful
     \item U1: Applies linear scaling
     \item \textit{Tag}: U1-improper-addition
     \end{itemize}
\end{itemize}

\textbf{12. Contextomy}
\begin{itemize}
     \item \textit{Definition}: Quoting out of context in a misleading way.
     \item \textit{Example}: "Time is money, so poorer countries have less time."
     \item \textit{KUDEX}:
     \begin{itemize}
     \item K1: Metaphor ("time is money")
     \item U1: Attends to monetary interpretation literally
     \item \textit{Tag}: U1-contextomy, A-on-term("money")
     \end{itemize}
\end{itemize}
}
\end{tcolorbox}
\caption{KUDEX-Fallacy Prompt For Fallacy Detection (Part 3)}
\label{fig:kudex-fallacy-prompt-3}
\end{figure}

\begin{figure}[h]
\centering
\begin{tcolorbox}[title=KUDEX-Fallacy Prompt For Fallacy Detection (continued)]
\small{
\textbf{13. Nominal Fallacy}
\begin{itemize}
     \item \textit{Definition}: Using a label as if it were an explanation.
     \item \textit{Example}: "He's angry because he has intermittent explosive disorder."
     \item \textit{KUDEX}:
     \begin{itemize}
     \item K1: Diagnostic label
     \item U1: Treats label as mechanism
     \item \textit{Tag}: U1-nominal-fallacy
     \item (Only a fallacy if no explanation accompanies the label)
     \end{itemize}
\end{itemize}

\textbf{14. Accident Fallacy}
\begin{itemize}
     \item \textit{Definition}: Misapplying a general rule to a special or exceptional case.
     \item \textit{Example}: "Speeding is illegal. So the ambulance driver should be arrested."
     \item \textit{KUDEX}:
     \begin{itemize}
     \item K1: General law
     \item U1: Applies to exceptional case (ambulance)
     \item D1: Infers criminality
     \item \textit{Tag}: U1-accident-fallacy or D1-accident-fallacy (depending on locus)
     \end{itemize}
\end{itemize}

\textbf{Final Note}:
Multiple tags may be applied to a single unit if multiple reasoning errors are present. Use your best judgment to distinguish where the fault lies — in the knowledge (K), the use (U), or the derivation (D). When in doubt, mark all plausible fallacies and be ready to explain the structural difference.

Your response should first analyze the statement using the KUDEX FALLACY DETECTION PROMPT, then summarize the fallacy types using XML format:

\begin{verbatim}
<verification_analysis>
[KUDEX FALLACY DETECTION PROMPT analysis output here]
</verification_analysis>
<fallacy_types>
    <!-- For every fallacy type: "yes" if it applies, otherwise "no" -->
    <FalseAnalogy>yes|no</FalseAnalogy>
    <Equivocation>yes|no</Equivocation>
    <FalsePremise>yes|no</FalsePremise>
    <WrongDirection>yes|no</WrongDirection>
    <FallacyOfComposition>yes|no</FallacyOfComposition>
    <BeggingTheQuestion>yes|no</BeggingTheQuestion>
    <FalseCause>yes|no</FalseCause>
    <InverseError>yes|no</InverseError>
    <ImproperTransposition>yes|no</ImproperTransposition>
    <ImproperDistributionOrAddition>yes|no</ImproperDistributionOrAddition>
    <Contextomy>yes|no</Contextomy>
    <NominalFallacy>yes|no</NominalFallacy>
    <AccidentFallacy>yes|no</AccidentFallacy>
    <FalseDilemma>yes|no</FalseDilemma>
</fallacy_types>
\end{verbatim}

Here is the example to analyze:
\begin{verbatim}
<example>
  <id>{{ id }}</id>
  <text>{{ text }}</text>
</example>
\end{verbatim}
}
\end{tcolorbox}
\caption{KUDEX-Fallacy Prompt For Fallacy Detection (Part 4)}
\label{fig:kudex-fallacy-prompt-4}
\end{figure}
\end{document}